\documentclass[]{fairmeta}

\usepackage{wrapfig}
\usepackage{tabularx}
\usepackage{textcomp}
\usepackage{stfloats}
\usepackage{url}
\usepackage{verbatim}
\usepackage{titlesec}
\usepackage{tocloft}
\usepackage{adjustbox}
\usepackage{multirow}
\usepackage{pifont}
\usepackage{tikz}
\usepackage{comment}
\usepackage{amsmath,amssymb} 
\usepackage{colortbl}  
\usepackage{color}
\usepackage{booktabs} 
\usepackage{hyperref}
\usepackage{graphicx}    
\usepackage{subcaption} 
\usepackage{multirow} 
\usepackage{booktabs} 
\usepackage{subcaption} 
\usepackage[table]{xcolor}
\RequirePackage{xspace}
\makeatletter
\DeclareRobustCommand\onedot{\futurelet\@let@token\@onedot}
\def\@onedot{\ifx\@let@token.\else.\null\fi\xspace}

\def\eg{\emph{e.g}\onedot}

\makeatother

\definecolor{adptorange}{RGB}{248, 205, 172}
\definecolor{cmpblue}{RGB}{189, 215, 238}
\definecolor{cmpblue}{RGB}{189, 215, 238}

\definecolor{our_red}{RGB}{232,157,160}
\definecolor{our_blue}{RGB}{136,206,230}
\definecolor{our_orange}{RGB}{246,200,168}
\definecolor{our_green}{RGB}{178,211,164}

\definecolor{attn_code0}{RGB}{247,215,200}
\definecolor{attn_code1}{RGB}{238,169,139}
\definecolor{mlp_code0}{RGB}{204,201,221}
\definecolor{mlp_code1}{RGB}{102,95,153}

\definecolor{token_blue}{RGB}{84, 120, 140}

\usepackage[table]{xcolor}
\usepackage{booktabs}
\usepackage{graphicx}

\definecolor{ftbg}{RGB}{246,245,252}
\definecolor{ftline}{RGB}{112,96,205}

\newcommand{\ftleft}[1]{%
    \multicolumn{1}{
        !{\color{ftline}\vrule width 0.8pt}c
    }{\cellcolor{ftbg}#1}%
}

\newcommand{\ftmid}[1]{%
    \cellcolor{ftbg}#1%
}

\newcommand{\plainleft}[1]{%
    \multicolumn{1}{
        !{\color{ftline}\vrule width 0.8pt}c
    }{#1}%
}

\usepackage{pifont}       
\usepackage{bbding}       
\usepackage{fontawesome}
\usepackage{xspace}

\usepackage{float}

\newlength\savewidth

\newcolumntype{x}[1]{>{\centering\arraybackslash}p{#1pt}}
\newcolumntype{y}[1]{>{\raggedright\arraybackslash}p{#1pt}}
\newcolumntype{z}[1]{>{\raggedleft\arraybackslash}p{#1pt}}

\renewcommand{\paragraph}[1]{\vspace{1mm}\noindent\textbf{#1}}
\usepackage{colortbl}
\usepackage{xcolor}
\usepackage{wrapfig}

\renewcommand{\paragraph}[1]{\vspace{1.25mm}\noindent\textbf{#1}}

\usepackage{algorithm}
\usepackage{listings}

\definecolor{codeblue}{rgb}{0.25, 0.5, 0.5}
\definecolor{codekw}{rgb}{0.35, 0.35, 0.75}
\lstdefinestyle{Pytorch}{
    language = Python,
    backgroundcolor = \color{white},
    basicstyle = \fontsize{9pt}{8pt}\selectfont\ttfamily\bfseries,
    columns = fullflexible,
    aboveskip=1pt,
    belowskip=1pt,
    breaklines = true,
    captionpos = b,
    commentstyle = \color{codeblue},
    keywordstyle = \color{codekw},
}

\definecolor{green}{HTML}{009000}
\definecolor{red}{HTML}{ea4335}

\title{RadSight: Towards Perceptually Reliable Multimodal Radiology Image Understanding}

\author[* 1, 2, 3]{Jianqin Liu}
\author[* 1, 2]{Weiwei Cao}
\author[* 1, 2]{Wanxing Chang}
\author[* 1, 2]{Ruifeng Yuan}
\author[* 1, 2]{Bowen Shi}
\author[1, 2]{Zhilin Zheng}
\author[1, 2]{Xianjie Zhang}
\author[1]{Ling Zhang}
\author[\dagger 3]{Peng Wang}
\author[\dagger 1, 2]{Jianpeng Zhang}

\affiliation[1]{DAMO Academy, Alibaba Group}
\affiliation[2]{Hupan Lab\\}
\affiliation[3]{University of Electronic Science and Technology of China}
\contribution[*]{Equal contribution}
\contribution[\dagger]{Corresponding author (Email: jianpeng.zhang0@gmail.com)}

\abstract{
    Medical multimodal large language models (MLLMs) are increasingly expected to perform complex image understanding tasks, yet their reliability is often compromised by frequent errors in visual interpretation. To systematically trace these failures, we traverse the hierarchy from high-level clinical tasks down to fundamental visual perception. We therefore introduce Perception-Bench, a large-scale benchmark comprising 1.13 million samples that assesses medical MLLMs across six dimensions: attribute judgment, spatial grounding, spatial understanding, disease prediction, anomaly detection, and report generation, spanning both 2D and 3D radiology images.
    Our analysis on Perception-Bench reveals that existing MLLMs lack the ability to capture even the most basic lesion attributes, such as location, size, and density. This inability to ground clinical outputs in primary visual evidence reveals that the models' diagnostic unreliability is rooted in a critical but overlooked bottleneck in low-level visual perception.
    Motivated by this, we propose RadSight, a perception-driven MLLM built upon a dual 2D/3D encoder architecture that preserves native imaging spatial structures. RadSight formulates medical image understanding as a four-stage progressive process: visual-language alignment, fine-grained visual perception, clinical diagnosis, and diagnostic interpretation. The model is trained on an 8.37 million perception-oriented corpus using progressive curriculum learning.
    On Perception-Bench, RadSight consistently outperforms existing MLLMs across all six evaluation dimensions, with particularly strong gains in spatial grounding and clinical diagnosis. It also achieves consistent improvements on public 2D and 3D medical benchmarks, further demonstrating that robust low-level visual perception is a critical foundation for reliable clinical understanding. Code and model will be publicly available.
}

\date{\today}
\metadata[Homepage]{
    \url{https://github.com/alibaba-damo-academy/damo-RadSight/}
}

\begin{document}
\thispagestyle{firstheader}
\maketitle
\pagestyle{empty}

\begin{figure*}[t]
  \centering
  \includegraphics[width=1\linewidth,clip=false, trim=0 8 5 0]{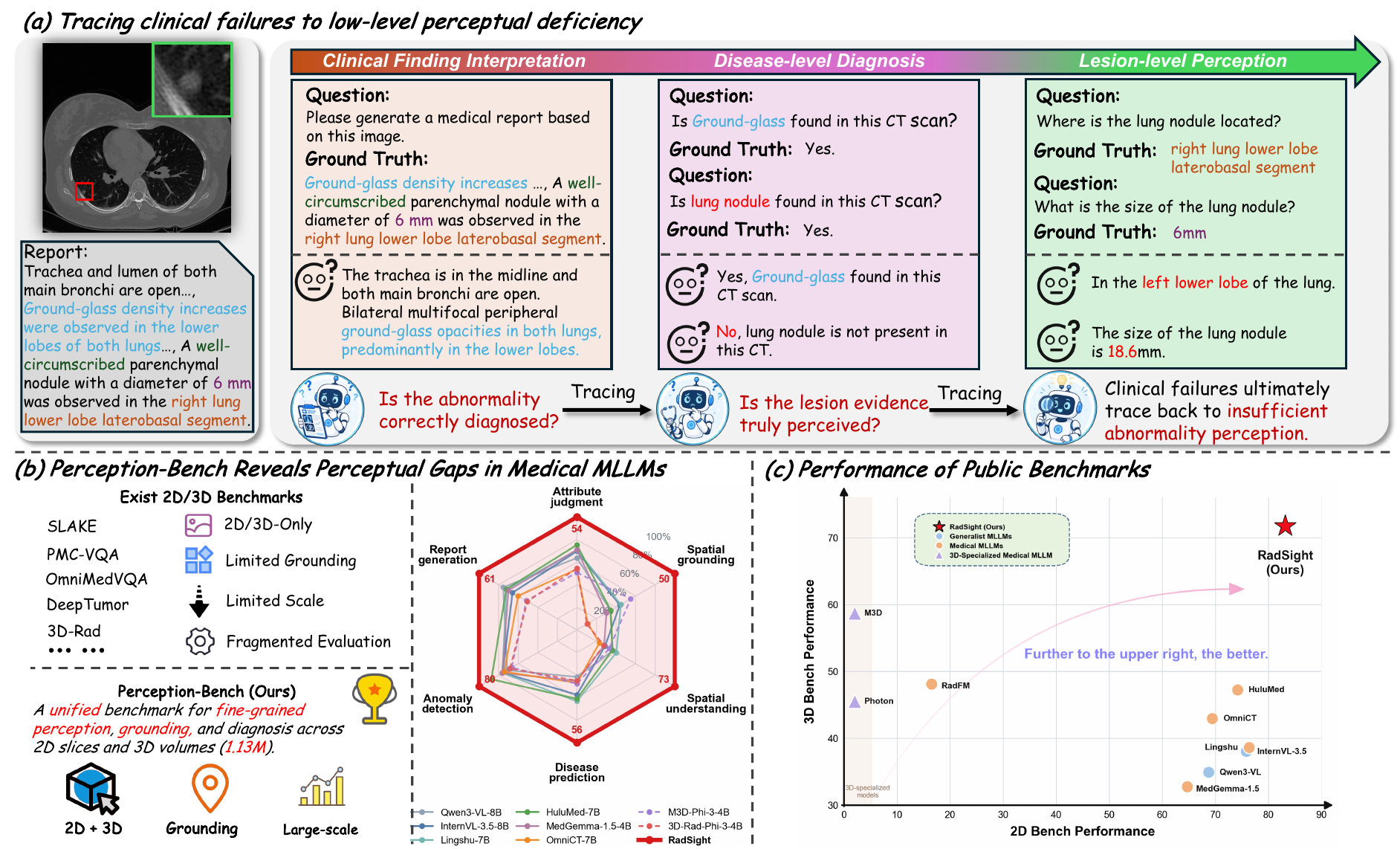}
    \vspace{-2mm}
    \caption{\textbf{Tracing clinical failures to perceptual limitations.} (a) A representative medical MLLM, Lingshu~\citep{xu2025lingshu} often misses abnormal findings in radiology reports. To trace the cause, we simplify the task from diagnosis to lesion attribute QA, yet the models still fail, indicating insufficient low-level visual perception. (b) Given that existing benchmarks cannot comprehensively and systematically evaluate low-level perceptual capabilities, we propose \textbf{Perception-Bench}, which reveals the perceptual deficiencies of current medical MLLMs. (c) RadSight outperforms the evaluated baselines on public medical benchmarks.
    }
  \label{fig:overview}
  \vspace{-4mm}
\end{figure*}

\section{Introduction} \label{sec:introduction}
Recent advances in large language models and vision-language modeling have accelerated the development of medical multimodal large language models (MLLMs)~\citep{moor2023foundation}. Representative systems, including Lingshu~\citep{xu2025lingshu}, Hulu-Med~\citep{jiang2025hulu}, and LLaVA-Med~\citep{li2023llava-med}, extend general-purpose vision-language models to biomedical and radiological settings. These systems are commonly developed through large-scale medical visual instruction tuning~\citep{li2023llava-med,zhang2023pmc}, visual-language alignment~\citep{boecking2022ms-cxr,li2023llava-med}, and supervision over diverse clinical tasks~\citep{tu2024medplam}. The resulting models support medical visual question answering~\citep{liu2021slake,zhang2023pmc}, abnormality recognition and diagnostic assistance~\citep{xu2025lingshu,jiang2025hulu}, and radiology report generation~\citep{chen2020g2gen,jin2024promptmrg,li2025towards}. Collectively, these capabilities position MLLMs as flexible interfaces for interpreting medical images and producing clinically relevant responses. 

Despite these advances, current medical MLLMs fall short when held to the standard of clinical reliability.
A common failure mode is that, when given a CT volume and asked to generate a radiology report, existing models often overlook critical abnormal findings or even produce overly normal reports~\citep{liu2024survey}, as shown in Figure~\ref{fig:overview}~(a). However, report generation is a high-level and sparse task: an incorrect report may arise from multiple sources, including weak language generation, insufficient diagnostic reasoning, or poor visual perception~\citep{ostmeier2024green, zhao2024ratescore, mahdavi2026med}. To trace the cause of these failures, we study this problem by treating task simplification as a diagnostic probe. We first convert report generation into clinical diagnosis by directly asking whether a specific disease is present. Surprisingly, existing models still produce incorrect answers. We then further reduce the task to lesion-level perception by asking about basic abnormality attributes, such as lesion location, size, and morphology. The models continue to fail on these simpler queries. This failure pattern reveals a critical bottleneck: 
\textbf{\emph{current medical MLLMs often struggle to perceive underlying visual abnormalities with sufficient clarity, thereby compromising their reliability across diagnosis, report generation, and more complex clinical tasks.}}

To comprehensively evaluate existing medical MLLMs from high-level clinical tasks down to fundamental visual perception, 
we introduce \textbf{Perception-Bench}, a large-scale perception-oriented benchmark for medical MLLMs. 
Existing medical benchmarks are often fragmented: 2D datasets~\citep{liu2021slake, hu2024omnimedvqa, zhang2023pmc, yamagishi2025radfig} primarily focus on slice-level VQA or classification-style question answering, while 3D benchmarks~\citep{3drad, chen2026deeptumorvqa, monon2026ct-spatial, yuan2026ct-finebench} mainly evaluate volumetric diagnosis or lesion-related reasoning. Despite their value, these benchmarks lack a unified, systematic framework to evaluate low-level visual perception across both 2D and 3D modalities.
Perception-Bench organizes evaluation as an evidence hierarchy with six dimensions: attribute judgment, spatial grounding, spatial understanding, disease prediction, anomaly detection, and report generation. Attribute judgment evaluates whether a model recognizes clinically meaningful properties of abnormalities. Spatial grounding measures concept-to-region localization, whereas spatial understanding tests the complementary region-to-concept mapping. Disease prediction and anomaly detection assess whether this evidence supports disease-level decisions, and report generation evaluates its integration into free-text clinical descriptions. The anomaly-detection component spans 82 disease categories across thoracic and abdominal anatomy. As illustrated in Figure~\ref{fig:overview}~(b), this comprehensive evaluation reveals that existing MLLMs suffer from substantial deficiencies in visual perception across multiple dimensions.

The deficiencies revealed by Perception-Bench motivate \textbf{RadSight}, a perception-driven medical MLLM designed around two principles. First, 2D medical images and 3D CT volumes should be encoded with modality-appropriate visual pathways. RadSight therefore uses a dedicated 2D image encoder and a dedicated 3D volumetric encoder, preserving in-plane visual detail and inter-slice anatomical context before projecting both representations into a shared language-model interface. Second, high-level clinical outputs should be learned after the model has acquired explicit evidence-level capabilities. We accordingly construct an 8.37-million-sample perception-oriented corpus and organize training into four progressive stages: visual-language alignment, fine-grained visual perception, clinical diagnosis, and report generation. This curriculum turns lesion attributes and spatial correspondence into explicit intermediate learning objectives instead of expecting them to emerge implicitly from diagnosis or report supervision.

Experiments on Perception-Bench show that RadSight improves performance across the full evaluation hierarchy relative to generalist, 2D medical, and 3D medical MLLMs. Evaluations on public 2D and 3D benchmarks further show that the learned representations transfer to external task formulations under the stated evaluation protocols. These results support our central hypothesis: \textbf{explicitly learning fine-grained visual evidence improves medical multimodal understanding on the evaluated tasks.}

Our contributions are summarized as follows:

$\bullet$ \textbf{Perception Failure Tracing}: We identify and analyze a fundamental limitation of current medical MLLMs: their high-level clinical failures often originate from insufficient low-level visual perception, as revealed by progressively simplifying tasks from report generation to diagnosis and lesion-attribute perception.

$\bullet$ \textbf{Perception-Bench}: We introduce Perception-Bench, a 1.13-million-sample benchmark that systematically evaluates medical MLLMs across six dimensions, covering fine-grained perception, spatial grounding, diagnosis, anomaly detection, and report generation over both 2D medical images and 3D CT volumes.

$\bullet$ \textbf{RadSight}: We develop a medical MLLM with dedicated 2D/3D visual encoders and a four-stage perception-driven curriculum trained on 8.37 million samples. Experiments show consistent improvements in perceptual, diagnostic, and report-generation performance on Perception-Bench and public medical benchmarks.

\section{Related Work}
\textbf{Medical Multimodal Large Language Models.}
Medical multimodal large language models (MLLMs) extend generative vision-language models to medical applications through domain-specific pretraining, vision-language alignment, and instruction tuning. Early systems such as LLaVA-Med~\citep{li2023llava-med} and Med-Flamingo~\citep{moor2023medflamingo} primarily established medical visual question answering and few-shot multimodal assistance over 2D biomedical images. More recent generalist models, including HealthGPT~\citep{lin2025healthgpt}, Hulu-Med~\citep{jiang2025hulu}, MedGemma~\citep{sellergren2026medgemma}, and Lingshu~\citep{xu2025lingshu}, broaden the task scope to medical image interpretation, diagnosis, and report generation. Much of this progress, however, has been developed around 2D image interfaces or slice-based representations, which do not explicitly preserve volumetric context when a CT study is represented by a limited set of slices. Volumetric MLLMs such as M3D~\citep{bai2024m3d}, CT-CHAT~\citep{hamamci2026ct-chat}, Merlin~\citep{merlin}, and Photon~\citep{fang2026photon} instead encode 3D scans and support tasks including volumetric question answering, diagnosis, localization, and report generation. RadFM~\citep{radfm} further provides a generalist interface for both 2D and 3D radiology data, while OmniCT~\citep{lin2026omnict} explicitly studies joint slice-volume CT understanding. These studies substantially expand the input and task coverage of medical MLLMs, but unified modeling alone does not ensure that high-level predictions are grounded in fine-grained visual evidence. RadSight therefore focuses on a complementary objective: it combines dedicated 2D and 3D encoders with a progressive curriculum that connects alignment, perception, clinical diagnosis, and interpretation.

\textbf{Medical Vision-Language Benchmarks.}
Medical vision-language benchmarks have evolved from task-specific 2D question answering toward broader multimodal and volumetric evaluation. VQA-RAD~\citep{lau2018vqa-rad}, SLAKE~\citep{liu2021slake}, and PathVQA~\citep{he2020pathvqa} evaluate question answering over radiology, semantically annotated medical images, and pathology images, respectively. PMC-VQA~\citep{zhang2023pmc}, OmniMedVQA~\citep{hu2024omnimedvqa}, and RadFig-VQA~\citep{yamagishi2025radfig} increase the scale or diversity of modalities and clinical questions. These benchmarks are valuable for measuring medical knowledge and image-level understanding, but they provide limited coverage of native volumetric perception and explicit bidirectional spatial correspondence. Recent 3D benchmarks address complementary aspects of this problem: M3D-Bench~\citep{bai2024m3d} evaluates multiple 3D medical tasks, 3D-RAD~\citep{3drad} and DeepTumorVQA~\citep{chen2026deeptumorvqa} emphasize volumetric and tumor-related reasoning, CT-SpatialVQA~\citep{monon2026ct-spatial} targets semantic-spatial understanding, and CT-FineBench~\citep{yuan2026ct-finebench} evaluates fine-grained factual consistency in CT reports. Taken together, prior benchmarks provide strong coverage of individual capabilities, but the benchmarks considered here do not jointly trace the full evidence hierarchy from low-level attributes and spatial correspondence to diagnosis and report generation across both native 2D and 3D inputs. Perception-Bench complements these resources with a unified six-dimensional evaluation of attribute judgment, spatial grounding, spatial understanding, disease prediction, anomaly detection, and report generation, thereby exposing where perceptual errors emerge and how they propagate to higher-level clinical outputs.
\section{Perception-Oriented Data Construction}

\begin{figure*}[t]
  \centering
  \includegraphics[width=1\linewidth,clip=false, trim=0 8 5 0]{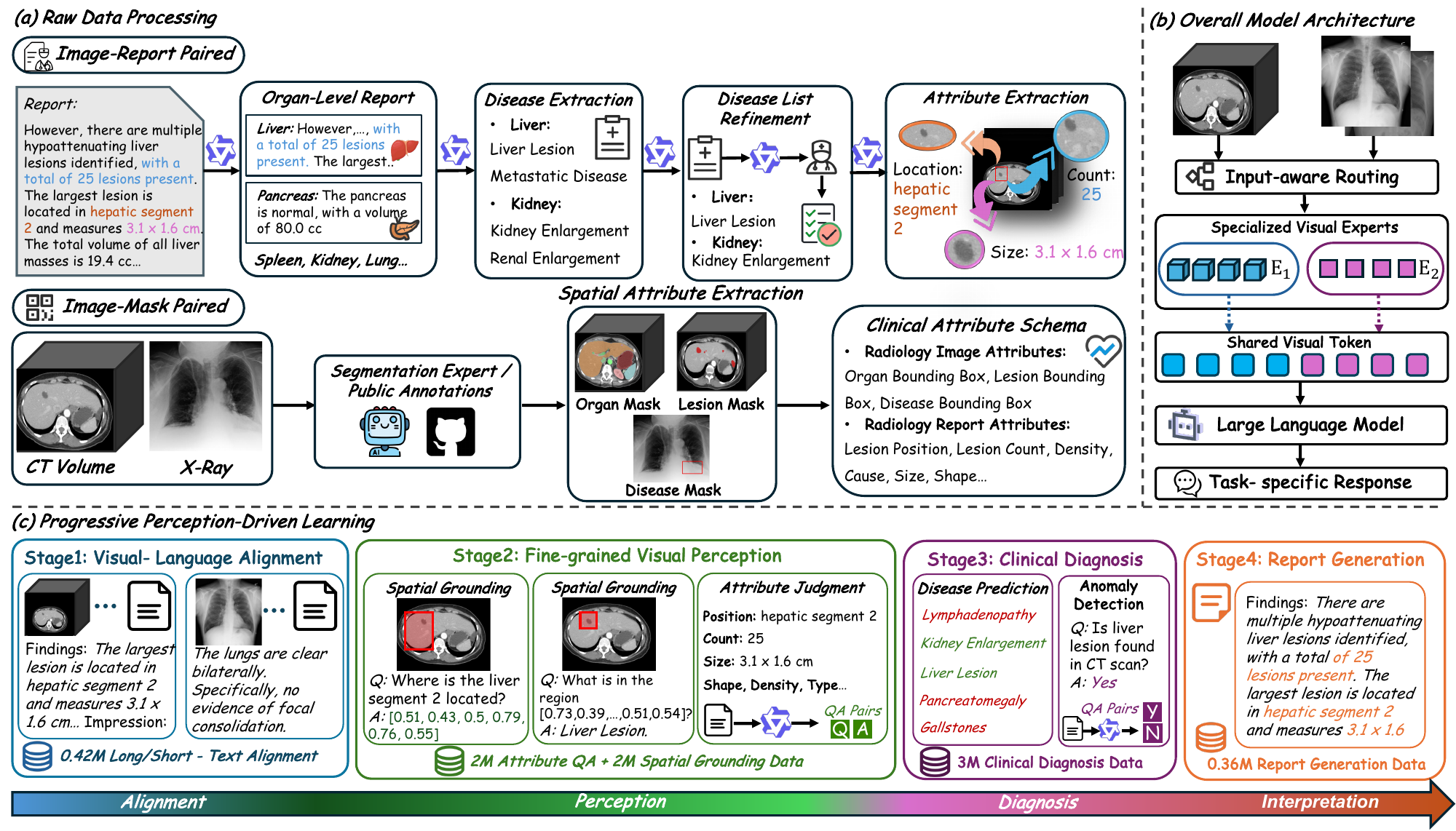}
    \vspace{-2mm}
    \caption{\textbf{(a) Raw Data Processing.}
    Heterogeneous image-report and image-mask pairs are converted into a unified clinical attribute schema, including report-derived lesion attributes and image-derived spatial annotations. \textbf{(b) Overall Model Architecture.} RadSight adopts a dual-encoder architecture to separately process 3D CT volumes and 2D medical images, with modality-specific projectors mapping visual features into the LLM embedding space for task-specific generation. \textbf{(c) Progressive Perception-Driven Learning.} RadSight trains through a four-stage curriculum, enabling high-level clinical output to be built on reliable visual perception.
    }
  \label{fig:data}
  \vspace{-3mm}
\end{figure*}
Perception-oriented evaluation and training require a common representation of the visual evidence underlying clinical outputs. Public medical datasets provide complementary signals, including radiology reports, organ masks, lesion annotations, and disease-level bounding boxes. These signals, however, are distributed across modalities and annotation formats and cannot be directly compared or combined. We therefore construct a unified annotation pipeline that converts heterogeneous 2D and 3D medical data into structured clinical concepts, lesion attributes, and spatial regions. As illustrated in Figure~\ref{fig:data}~(a), the pipeline contains two parallel streams: report-derived semantic annotation and image-derived spatial annotation.

\subsection{Data Source.}
We curate two complementary families of public datasets. The first contains medical images paired with radiology reports. For 2D chest X-rays, we use MIMIC-CXR~\citep{mimic} and IU-Xray~\citep{iu_xray}. For 3D CT, we use report-associated data from CT-RATE~\citep{ct-rate}, Merlin~\citep{merlin}, INSPECT~\citep{inspect}, AMOS-MM~\citep{amos}, and AbdomenAtlas~3.0~\citep{radgpt}, together with additional low-dose CT volumes from NLST~\citep{nlst}. The paired reports provide semantic supervision for disease extraction, lesion-attribute construction, visual-language alignment, and report generation.

The second family provides spatial annotations. PadChest-GR~\citep{de2025padchest_gr}, VinDr-CXR~\citep{nguyen2022vindr}, MS-CXR~\citep{boecking2022ms-cxr}, and REFLACX~\citep{bigolin2022reflacx} provide disease- or region-level annotations for chest X-rays. For CT volumes, we apply TotalSegmentator~\citep{wasserthal2023totalsegmentator} to CT-RATE, Merlin, and AbdomenAtlas~3.0 to obtain model-derived organ masks. We further use dataset-provided lesion and nodule annotations from AbdomenAtlas~3.0. A complete summary of the data sources is provided in Appendix~\ref{sec:appendix-data_sources}.

\subsection{Unified Annotation Construction}
\label{sec:unified_annotation}

\paragraph{Report-derived semantic annotations.}
Free-text reports describe clinically relevant findings but do not provide a consistent representation of organs, diseases, and lesion attributes. We convert each report into structured annotations using a constrained two-step pipeline. First, Qwen3-Max~\citep{bai2025qwen3} decomposes the report into organ-level sub-reports and associates each organ with its supporting sentences. It then extracts diseases or imaging findings together with their textual evidence and status (\emph{present}, \emph{absent}, \emph{suspected}, or \emph{historical}). Synonymous expressions are normalized into a common disease vocabulary, followed by label filtering and manual review of the resulting vocabulary. Second, for each retained disease or imaging finding, the model extracts only attributes explicitly stated in the supporting report text. The candidate schema includes location, size, count, shape, density, margin, type, and other report-supported properties. Each attribute--value pair is converted into a question--answer instance. The prompt prohibits unsupported inference and requires every target answer to be directly traceable to the source report. Prompt templates and output constraints are provided in Appendix~\ref{sec:appendix-report_parsing}.

\paragraph{Image-derived spatial annotations.}
We convert segmentation masks and native bounding boxes into normalized spatial regions. Each 2D region is represented by four corner coordinates, $(x_1,y_1,x_2,y_2)$, whereas each 3D region is represented by six coordinates, $(x_1,y_1,z_1,x_2,y_2,z_2)$. All coordinates are normalized to $[0,1]$ relative to the processed image or volume dimensions. For chest X-rays, regions associated with the same disease are grouped and duplicate boxes are removed. For CT, organ boxes are obtained from the tight spatial extents of TotalSegmentator masks, whereas disconnected lesion components are processed separately to preserve instance-level lesion boxes. Detailed resampling, intensity normalization, component filtering, and box extraction procedures are provided in Appendix~\ref{sec:appendix-data_process}.

The resulting schema supports six task definitions: attribute judgment, spatial grounding, spatial understanding, disease prediction, anomaly detection, and report generation. We use these shared definitions to construct two products with different roles: Perception-Bench for held-out evaluation and a separate perception-oriented corpus for training RadSight.

\subsection{Perception-Bench}
\label{sec:perception_bench}
Perception-Bench evaluates whether a medical MLLM can recover the visual evidence required for progressively higher-level clinical outputs. Rather than treating medical image understanding as a single end task, the benchmark organizes evaluation along an evidence hierarchy. It first measures fine-grained attributes and spatial correspondence, then evaluates disease-level decisions, and finally assesses their integration into a radiology report. This organization enables high-level errors to be traced to more specific perceptual or diagnostic failures without assuming that perception is the only source of clinical error. Detailed data distributions are provided in
Appendix~\ref{sec:appendix-perception_bench_dist}.

\paragraph{Fine-grained perception.}
Perception-Bench includes three tasks that directly examine visual evidence in an input $X^{m}$, where $m\in\{\mathrm{2D},\mathrm{3D}\}$. \emph{Attribute judgment} predicts a set of clinically meaningful properties $\widehat{\mathcal{A}}_{c}$ for a lesion or abnormality $c$, including its location, size, count, density, margin, and morphology. \emph{Spatial grounding} evaluates concept-to-region correspondence by mapping a specified organ, lesion, or disease concept $c$ to its normalized region $\widetilde{\mathbf{b}}^{m}$. \emph{Spatial understanding} evaluates the inverse mapping, $\widetilde{\mathbf{b}}^{m}\mapsto c$, by requiring the model to identify the anatomical or pathological concept associated with a given 2D or 3D region. Together, these tasks assess what visual evidence the model perceives and where that evidence is located.

\paragraph{Clinical diagnosis.}
The next level evaluates whether fine-grained evidence supports disease-level decisions. Given a candidate set $\mathcal{C}=\{c_1,\ldots,c_K\}$, \emph{disease prediction} outputs the subset of abnormalities present in the input, denoted by $\widehat{\mathcal{D}}\subseteq\mathcal{C}$. In contrast, \emph{anomaly detection} performs binary verification for a queried abnormality $c$ and predicts $\widehat{a}_{c}\in\{\mathrm{Yes},\mathrm{No}\}$. Disease prediction measures multi-label clinical recognition, whereas anomaly detection supports disease-wise evaluation using model confidence scores.

\paragraph{Report generation.}
The final task evaluates the integration of visual and diagnostic evidence into a free-text radiology report. Given an input $X^{m}$, the model generates a report $\mathbf{Y}^{\mathrm{rep}}=(y_1^{\mathrm{rep}},\ldots,y_{N_{\mathrm{rep}}}^{\mathrm{rep}})$ containing the corresponding findings and impression. Because an incorrect report may arise from perceptual, diagnostic, or language-generation errors, report generation is evaluated alongside the preceding evidence-level tasks rather than interpreted in isolation.

\begin{figure*}[t]
  \centering
  \includegraphics[width=1\linewidth,clip=false, trim=0 8 5 0]{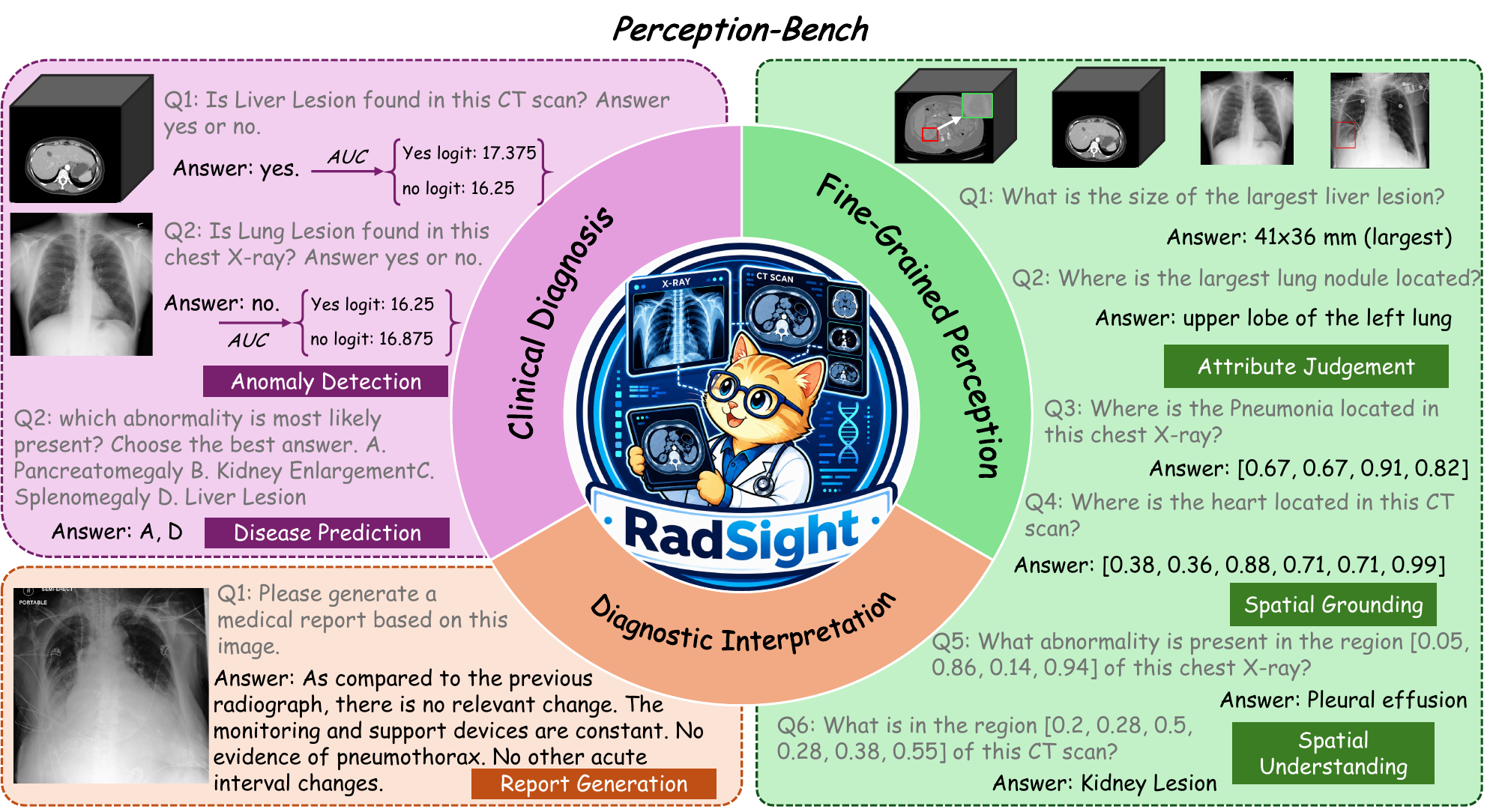}
    \vspace{-2mm}
    \caption{\textbf{Task hierarchy of Perception-Bench.}
    The benchmark evaluates six dimensions: attribute judgment, spatial grounding, spatial understanding, disease prediction, anomaly detection, and report generation.
    }
  \label{fig:data_dist}
  \vspace{-3mm}
\end{figure*}

\section{RadSight}
RadSight is a perception-driven medical multimodal large language model designed to process both 2D medical images and 3D CT volumes within a shared generative interface. Its design addresses two complementary requirements. First, 2D images and volumetric CT scans have different spatial structures and therefore require modality-specific visual encoding. Second, high-level clinical outputs should be learned after the model has acquired explicit supervision for fine-grained attributes and spatial evidence. As illustrated in Figure~\ref{fig:data}~(b--c), RadSight combines a dual-encoder visual architecture with a four-stage curriculum that progresses from visual-language alignment to fine-grained perception, clinical diagnosis, and  diagnostic interpretation.

\subsection{Unified Medical Visual Encoding.}
RadSight uses separate visual pathways for 2D medical images and 3D CT volumes. The 2D pathway captures in-plane appearance and spatial organization, whereas the 3D pathway jointly models volumetric anatomy and inter-slice context. This design avoids decomposing a CT volume into independent slices, which can discard information about lesion extent, anatomical continuity, and spatial relationships across slices.

Let $X^{\mathrm{2D}}\in\mathbb{R}^{H\times W\times C}$ denote a 2D medical image and $X^{\mathrm{3D}}\in\mathbb{R}^{D\times H\times W\times C}$ denote a CT volume with $D$ slices. For modality $m\in\{\mathrm{2D},\mathrm{3D}\}$, RadSight routes $X^m$ through the corresponding visual encoder:
\begin{equation}
F^m=E^m(X^m)\in\mathbb{R}^{N_m\times d_m^{\mathrm{enc}}},
\qquad
m\in\{\mathrm{2D},\mathrm{3D}\},
\label{eq:modality_encoding}
\end{equation}
where $E^m$ is the modality-specific encoder, $N_m$ is the number of output visual tokens, and $d_m^{\mathrm{enc}}$ is the encoder feature dimension. We instantiate $E^{\mathrm{2D}}$ with the SigLIP-NaViT encoder from VideoLLaMA3~\citep{zhang2025videollama}. For $E^{\mathrm{3D}}$, we use a PlainConvUNet-based encoder from nnU-Net~\citep{isensee2021nnu}, initialized from a pretrained 3D segmentation model. The deepest 3D feature map is projected with a $1\times1\times1$ convolution and flattened into a sequence of volumetric visual tokens.

Because the two encoders produce features with different dimensions and distributions, RadSight employs a modality-specific projector $P^m$ for each pathway:
\begin{equation}
Z^m=P^m(F^m)\in\mathbb{R}^{N_m\times d_{\mathrm{LLM}}},
\label{eq:visual_projection}
\end{equation}
where $P^m$ is a two-layer multilayer perceptron and $d_{\mathrm{LLM}}$ is the embedding dimension of the language model. The projected tokens preserve modality-specific visual evidence while sharing the dimensionality required by the language backbone.

We use Qwen3-VL~\citep{bai2025qwen3} as the language backbone. Given an instruction token sequence $\mathbf{x}=(x_1,\ldots,x_{N_x})$ with embeddings $T=\operatorname{Emb}(\mathbf{x})$, the visual and textual tokens are concatenated as
\begin{equation}
H^m=[Z^m;T].
\label{eq:multimodal_context}
\end{equation}
The model then generates a target response $\mathbf{y}=(y_1,\ldots,y_{N_y})$ autoregressively:
\begin{equation}
p_{\Theta}(\mathbf{y}\mid X^m,\mathbf{x})
=
\prod_{t=1}^{N_y}
p_{\Theta}(y_t\mid y_{<t},H^m),
\label{eq:multimodal_generation}
\end{equation}
where $[\cdot\,;\cdot]$ denotes sequence concatenation and $\Theta$ denotes the model parameters. This shared interface supports textual answers, categorical decisions, spatial coordinates, and radiology reports without task-specific prediction heads.

\subsection{Progressive Perception-Driven Learning.}
\label{sec:progressive_learning}
Using the unified annotations described in Section~\ref{sec:unified_annotation}, we organize the RadSight training data into four stages: visual-language alignment, fine-grained visual perception, clinical diagnosis, and diagnostic interpretation. Each stage is initialized from the optimized checkpoint of the preceding stage. Unless otherwise specified, the sample counts below describe the newly introduced data at each stage before replay. Detailed modality distributions and effective training sizes after replay are provided in Appendix~\ref{sec:appendix-training_details}.

\textbf{Stage 1: Visual-Language Alignment.}
In the first stage, we perform visual-language alignment using medical images paired with both full-length radiology reports and concise descriptions of imaging findings. This stage aims to align the outputs of the visual encoders with the language embedding space of the LLM. During this stage, we freeze the LLM and optimize only the 2D and 3D vision encoders and their corresponding projectors. This alignment-focused training encourages the visual modules to produce tokens that are compatible with the LLM embedding space while preserving the pretrained linguistic and medical knowledge encoded in the LLM.

\textbf{Stage 2: Fine-Grained Visual Perception.}
The second stage introduces explicit supervision for visual attributes and spatial correspondence. Using the report-derived attributes and spatial annotations from Section~\ref{sec:unified_annotation}, we construct approximately 2.19M attribute judgment, 1.19M spatial grounding, and 1.18M spatial understanding instances, for a total of 4.56M samples. Attribute judgment requires RadSight to identify properties such as location, size, count, density, margin, and morphology. Spatial grounding and spatial understanding establish complementary mappings between clinical concepts and image regions:
\begin{equation}
\underbrace{
\mathcal{F}_{\Theta}
\left(X^m,\mathbf{x}_c\right)
\rightarrow\widetilde{\mathbf{b}}^m
}_{\text{spatial grounding}},
\qquad
\underbrace{
\mathcal{F}_{\Theta}
\left(X^m,\mathbf{x}_{\widetilde{\mathbf{b}}}\right)
\rightarrow c
}_{\text{spatial understanding}},
\label{eq:bidirectional_spatial_learning}
\end{equation}
where $c$ denotes an organ, lesion, or pathological concept, $\mathbf{x}_c$ is a concept-conditioned localization instruction, and $\mathbf{x}_{\widetilde{\mathbf{b}}}$ contains a specified spatial region. The target $\widetilde{\mathbf{b}}^m$ denotes a normalized 2D or 3D bounding box. These tasks provide direct supervision for recognizing visual properties and associating clinical concepts with localized evidence.

\textbf{Stage 3: Clinical Diagnosis.}
The third stage connects the evidence-level representations learned in Stage~2 with disease-level clinical semantics. We construct approximately 0.95M disease-prediction and 2.05M anomaly-detection instances, for a total of 3.00M newly introduced samples. Disease prediction uses the curated disease lists derived from radiology reports as positive labels, while negative candidates are sampled from absent or clinically plausible findings. Given a candidate set $\mathcal{C}=\{c_1,\ldots,c_K\}$, the model predicts the subset of present abnormalities $\widehat{\mathcal{D}}\subseteq\mathcal{C}$. Anomaly detection instead verifies a single queried abnormality $c$ and produces $\widehat{a}_c\in\{\mathrm{Yes},\mathrm{No}\}$. The two tasks can be summarized as:
\begin{equation}
\underbrace{
\mathcal{F}_{\Theta}
\left(X^m,\mathbf{x}_{\mathrm{dis}},\mathcal{C}\right)
\rightarrow\widehat{\mathcal{D}}
}_{\text{disease prediction}},
\qquad
\underbrace{
\mathcal{F}_{\Theta}
\left(X^m,\mathbf{x}_{\mathrm{ano}}(c)\right)
\rightarrow\widehat{a}_c
}_{\text{anomaly detection}}.
\label{eq:diagnostic_tasks}
\end{equation}
where $\widehat{\mathcal{D}}$ denotes the predicted diseases or imaging
findings and $\widehat{a}_{c}$ indicates whether the queried
abnormality $c$ is present. These tasks encourage RadSight to derive
diagnostic conclusions from observable visual attributes and localized
regions rather than relying predominantly on language priors.

\textbf{Stage 4: Diagnostic Interpretation.}
The final stage introduces approximately 363K image-to-report instances constructed from paired 2D images, 3D CT volumes, and radiology reports. Given an input $X^m$ and a report-generation instruction $\mathbf{x}_{\mathrm{rep}}$, RadSight generates
\begin{equation}
\mathcal{F}_{\Theta}
\left(X^m,\mathbf{x}_{\mathrm{rep}}\right)
\rightarrow
\mathbf{Y}^{\mathrm{rep}}
=
\left(y_1^{\mathrm{rep}},\ldots,y_{N_{\mathrm{rep}}}^{\mathrm{rep}}\right),
\label{eq:report_generation_task}
\end{equation}
where $\mathbf{Y}^{\mathrm{rep}}$ contains the generated findings and impression. By introducing report generation after explicit perception and diagnosis supervision, the curriculum encourages report-level outputs to draw on lesion attributes, localized regions, and disease-level semantics acquired during the preceding stages.

\subsection{Training Objectives.}
\label{sec:training_objectives}

All tasks in RadSight are formulated as instruction-following
generation and optimized using a unified autoregressive language
modeling objective. At stage $s$, let $\mathcal{D}_s$ denote the
corresponding training data, $H_m=[Z^m;T]$ the multimodal input
sequence, and $\mathbf{y}=(y_1,\ldots,y_{N_y})$ the target response.
The training objective is defined as: 
\begin{equation}
\mathcal{L}_{s}
=
-\mathbb{E}_{(H_m,\mathbf{y})\sim\mathcal{D}_s}
\left[
\frac{1}{N_y}
\sum_{t=1}^{N_y}
\log p_{\Theta_s}
\left(
y_t\mid H_m,y_{<t}
\right)
\right],
\qquad
s\in\{1,2,3,4\},
\label{eq:training_objective}
\end{equation}
where $\Theta_s$ denotes the model parameters at stage $s$. The loss is
computed only over the target response tokens, while the visual and
instruction tokens serve as conditioning context. Textual
descriptions, categorical predictions, binary answers, and normalized
bounding-box coordinates are therefore supervised under the same
generative objective without introducing task-specific prediction
heads. Training proceeds sequentially across the four stages:
\begin{equation}
\Theta_s^{(0)}
\leftarrow
\Theta_{s-1}^{*},
\qquad
\Theta_s^{*}
=
\underset{\Theta_s^{\mathrm{train}}}{\arg\min}\,
\mathcal{L}_{s},
\qquad
s\in\{2,3,4\},
\label{eq:progressive_optimization}
\end{equation}
where $\Theta_s^{(0)}$ is initialized from the optimized checkpoint
$\Theta_{s-1}^{*}$ of the preceding stage. During Stage~1, the LLM is
frozen and only the vision encoders and projectors are optimized. From
Stage~2 onward, all model components are jointly optimized.
\begin{table*}[t]
\vspace{-0.5cm}
\caption{Performance comparison on our perception-oriented medical understanding benchmark. We evaluate RadSight against generalist, 2D medical, and 3D medical MLLMs across attribute judgment, spatial grounding, spatial understanding, disease prediction, anomaly detection, and report generation. \textbf{Bold} and \underline{underlined} text indicates the best and second-best performance, respectively.}
\vspace{-1mm}
\label{tab:tumor_bench}
\begin{center}
\renewcommand{\arraystretch}{1.2} 
\resizebox{\linewidth}{!}{
\begin{tabular}{l c c c c c c c c c c c} 
\toprule
 & \multicolumn{1}{c}{\textbf{Attribute Judgment}} & \multicolumn{2}{c}{\textbf{Spatial Grounding}} & \multicolumn{2}{c}{\textbf{Spatial Understanding}} & \multicolumn{1}{c}{\textbf{Disease Prediction}} & \multicolumn{3}{c}{\textbf{Anomaly Detection}} & \multicolumn{2}{c}{\textbf{Report}} \\
\cmidrule(lr){2-2} \cmidrule(lr){3-4} \cmidrule(lr){5-6} \cmidrule(lr){7-7} \cmidrule(lr){8-10} \cmidrule(lr){11-12}
\textbf{Model} & \textbf{LLM Score} & \textbf{2D-mIoU} & \textbf{3D-mIoU} & \begin{tabular}{@{}c@{}}\textbf{Disease}\\\textbf{Pos./2D}\end{tabular} & \begin{tabular}{@{}c@{}}\textbf{Organ/Lesion}\\\textbf{Pos./3D}\end{tabular} & \textbf{Accuracy} & \textbf{AUC} & \textbf{Sens} & \textbf{Spec} & \textbf{RaTEScore} & \textbf{CT-Rate-F1} \\
\midrule

\rowcolor{gray!5}
\multicolumn{12}{c}{\textit{Generalist MLLMs}} \\
\midrule
Qwen3-VL-8B & 34.54 & 17.14 & -- & 23.35 & 30.54 & 23.27 & 57.36 & 53.46 & 63.45 & 46.19 & 9.39 \\
InternVL-3.5-8B & 37.77 & 21.86 & -- & 10.60 & 35.81 & 32.07 & 61.10 & 64.81 & 55.14 & 40.36 & 3.96 \\
\midrule

\rowcolor{gray!5}
\multicolumn{12}{c}{\textit{2D Medical MLLMs}} \\
\midrule
Lingshu-7B & 38.26 & 22.36 & -- & 22.07 & 37.24 & 35.31 & 61.40 & 66.23 & 54.80 & 44.04 & 7.66 \\
HuluMed-7B & 40.81 & 17.17 & -- & 26.56 & 26.56 & 34.10 & 70.39 & 69.01 & 64.09 & 44.23 & 23.92 \\
MedGemma-1.5-4B & 38.46 & 15.02 & -- & 20.04 & 18.25 & 25.26 & 61.37 & 67.18 & 55.35 & 42.88 & 11.62 \\
OmniCT-7B & 29.15 & 5.54 & -- & 13.70 & 20.14 & 26.21 & 59.72 & 63.20 & 54.28 & 36.82 & 10.19 \\
\midrule

\rowcolor{gray!5}
\multicolumn{12}{c}{\textit{3D Medical MLLMs}} \\
\midrule
M3D-Phi-3-4B & 27.47 & -- & 27.36 & -- & 24.29 & 26.64 & 53.63 & 69.10 & 41.50 & 31.65 & 3.59 \\
3D-Rad-Phi-3-4B & 29.54 & -- & 5.15 & -- & 24.01 & 25.12 & 55.09 & 62.67 & 50.30 & 30.98 & 6.11 \\
\midrule

\rowcolor{gray!5}
\multicolumn{12}{c}{\textit{Our Models}} \\
\midrule
\rowcolor{blue!5} \textbf{RadSight-4B} & \underline{52.45} & \underline{34.19} & \textbf{64.68} & \underline{48.25} & \underline{96.63} & \textbf{56.16} & \textbf{81.10} & \textbf{78.02} & \textbf{72.35} & \underline{60.73} & \textbf{36.45} \\
\rowcolor{blue!5} \textbf{RadSight-8B} & \textbf{54.15} & \textbf{35.20} & \underline{64.31} & \textbf{50.11} & \textbf{96.64} & \underline{55.81} & \underline{80.47} & \underline{77.19} & \underline{72.10} & \textbf{61.24} & \underline{36.00} \\
\bottomrule
\end{tabular}
}
\end{center}
\end{table*}

\section{Experiments}
\subsection{Experimental Settings}
\textbf{Implementation Details.}
We build RadSight on two pretrained LLM backbones, Qwen3-VL-4B and Qwen3-VL-8B, resulting in RadSight-4B and RadSight-8B, respectively. RadSight adopts two modality-specific vision encoders. Specifically, we use SigLIP~\citep{zhang2025videollama} as the 2D vision encoder to extract visual representations from slice-based medical images. For 3D CT volumes, we use Radar~\citep{zhang2026radar} as the volumetric vision encoder to capture 3D anatomical structures and inter-slice spatial context. For public 2D and 3D benchmark evaluation, we further fine-tune RadSight, after the four-stage perception-oriented training, on the combined training splits of all public benchmarks. The same fine-tuned checkpoint is evaluated on the test split of each benchmark. This setting evaluates the transferability of RadSight’s perception-oriented representation to external medical benchmarks. Additional details on data processing, model architecture, public benchmark protocols, and training configurations are provided in
Appendix~\ref{sec:appendix-implementation}.

\textbf{Baseline Methods.}
We compare RadSight with three groups of representative baselines. (1) Generalist Models, including Qwen3-VL-8B~\citep{bai2025qwen3} and InternVL-3.5-8B~\citep{wang2025internvl3}. (2) 2D Medical Models, including Lingshu-7B~\citep{xu2025lingshu}, HuluMed-7B~\citep{jiang2025hulu}, MedGemma-1.5-4B~\citep{sellergren2026medgemma} and OmniCT-7B~\citep{lin2026omnict}. (3) 3D Medical Models, including RadFM~\citep{radfm}, Photon~\citep{fang2026photon}, M3D-Phi-3-4B~\citep{bai2024m3d} and 3D-Rad-Phi-3-4B~\citep{3drad}.

\begin{table*}[t]
\caption{Comparison with existing MLLMs on public 2D medical benchmarks. \textbf{Bold} and
\underline{underlined} values indicate the best and second-best
results, respectively.}
\label{tab:2d_benchmark}
\begin{center}
\renewcommand{\arraystretch}{1.0}
\resizebox{0.85\linewidth}{!}{
\begin{tabular}{l c c c c c c c c} 
\toprule
\multirow{2}{*}{\textbf{Model}} & \multicolumn{2}{c}{\textbf{SLAKE}} & \multicolumn{1}{c}{\textbf{PMC-VQA}} & \multicolumn{1}{c}{\textbf{OmniMedVQA}} & \multicolumn{3}{c}{\textbf{RadFig-VQA}} & \multirow{2}{*}{\textbf{Avg}} \\
\cmidrule(lr){2-3} \cmidrule(lr){4-4} \cmidrule(lr){5-5} \cmidrule(lr){6-8}
 & \textbf{Open} & \textbf{Closed} & \textbf{Choice} & \textbf{Choice} & \textbf{Easy} & \textbf{Medium} & \textbf{Hard} & \\
\midrule

\rowcolor{gray!5}
\multicolumn{9}{c}{\textit{Generalist MLLMs}} \\ 
\midrule
Qwen3-VL-8B & 65.50 & 74.76 & 46.05 & 75.96 & 74.48 & 72.92 & 71.26 & 68.70 \\
InternVL-3.5-8B & 71.38 & 78.59 & 54.72 & 89.21 & 78.56 & 79.09 & 79.27 & 75.83 \\
\midrule

\rowcolor{gray!5}
\multicolumn{9}{c}{\textit{Medical MLLMs}} \\ 
\midrule
RadFM-14B & 31.73 & 18.58 & 14.46 & 16.28 & 16.55 & 8.67 & 9.29 & 16.51 \\
Lingshu-7B & 78.22 & 81.94 & 55.15 & 82.12 & 80.56 & 77.66 & 78.82 & 76.35 \\
HuluMed-7B & 66.45 & 76.08 & \underline{60.95} & 81.13 & 77.67 & 78.02 & 78.99 & 74.18 \\
MedGemma-1.5-4B & 75.84 & 68.28 & 47.89 & 69.81 & 64.51 & 62.88 & 63.71 & 64.70 \\
OmniCT-7B & 59.70 & 69.62 & 49.36 & 79.25 & 75.27 & 76.31 & 76.57 & 69.44 \\
\midrule

\rowcolor{gray!5}
\multicolumn{9}{c}{\textit{Ours}} \\ 
\midrule
\rowcolor{blue!5} \textbf{RadSight-4B} & \underline{80.21} & \underline{85.05} & 60.81 & \underline{99.39} & \underline{82.05} & \textbf{81.79} & \underline{83.23} & \underline{81.79} \\
\rowcolor{blue!5} \textbf{RadSight-8B} & \textbf{82.27} & \textbf{88.16} & \textbf{62.30} & \textbf{99.51} & \textbf{84.05} & \underline{81.75} & \textbf{84.24} & \textbf{83.18} \\
\bottomrule
\end{tabular}
}
\end{center}
\vspace{-0.5cm}
\end{table*}

\subsection{Evaluation Metrics}
\paragraph{Attribute judgment.}
We evaluate open-ended attribute responses using Qwen3-Max as an automatic judge. For each instance, the evaluator compares the prediction with the reference answer and assigns a score of $0$, $0.5$, or $1$, corresponding to incorrect, partially correct, and fully correct responses. Categorical answers are assessed for semantic equivalence and completeness. For quantitative attributes, units are first normalized, after which relative errors of at most 10\%, between 10\% and 30\%, and above 30\% receive scores of $1$, $0.5$, and $0$, respectively. We average the instance-level scores and report the result as a percentage. The complete scoring rubric and evaluation prompt are provided in Appendix~\ref{sec:appendix-attribute_eval_prompt}.

\paragraph{Spatial grounding and spatial understanding.}
For spatial grounding, we compute the intersection over union between predicted and reference boxes and report 2D and 3D mean IoU (mIoU) separately. When an instance contains multiple regions, predicted and reference boxes are matched using the procedure described in Appendix~\ref{sec:appendix-grounding_metrics}. For spatial understanding, we report recognition accuracy separately for 2D disease regions and 3D organ or lesion regions.

\paragraph{Clinical diagnosis.}
For anomaly detection, threshold-dependent accuracy can be misleading when positive and negative cases are imbalanced. We therefore use the macro-averaged area under the receiver operating characteristic curve (AUROC) as the primary metric. For each queried clinical category, we use the model probability assigned to the positive response as the prediction score, compute AUROC within the corresponding imaging modality, and then average across categories. We additionally report sensitivity and specificity at the operating threshold selected using Youden's index~\citep{youden1950index}. The complete diagnostic evaluation protocol is provided in Appendix~\ref{sec:appendix-clinical_metrics}.

\paragraph{Report generation.}
We evaluate generated reports using RaTEScore~\citep{zhao2024ratescore} and Clinical Efficacy (CE) F1~\citep{hamamci2026ct-chat}. Detailed metrics are provided in Appendix~\ref{sec:appendix-mrg_metrics}.
\begin{table*}[t]
\caption{
Comparison of RadSight with existing MLLMs on public 3D CT benchmarks.
Light-purple blocks indicate benchmarks whose training data were used
to train or fine-tune the corresponding model. \textbf{Bold} and
\underline{underlined} values indicate the best and second-best
results, respectively.
}
\label{tab:3d_benchmark}
\centering

\renewcommand{\arraystretch}{1.2}

\resizebox{\linewidth}{!}{
\begin{tabular}{l c c c c c c c c c}

\toprule

\textbf{Model}
& \multicolumn{3}{c}{\textbf{3D-Rad}}
& \multicolumn{5}{c}{\textbf{DeepTumorVQA}}
& \textbf{CT-Spatial-VQA} \\

\cmidrule(lr){2-4}
\cmidrule(lr){5-9}
\cmidrule(lr){10-10}

&
\begin{tabular}{@{}c@{}}
\textbf{Existence}\\
\textbf{Detection}
\end{tabular}
&
\begin{tabular}{@{}c@{}}
\textbf{Static Temporal}\\
\textbf{Diagnosis}
\end{tabular}
&
\begin{tabular}{@{}c@{}}
\textbf{Longitudinal}\\
\textbf{Temporal Diagnosis}
\end{tabular}
&
\textbf{Recog.}
&
\textbf{Meas.}
&
\textbf{Vis.R.}
&
\textbf{Med.R.}
&
\textbf{Overall}
&
\textbf{Accuracy} \\

\midrule


\rowcolor{gray!5}
\multicolumn{10}{c}{\textit{Generalist MLLMs}} \\

\midrule

Qwen3-VL-8B
& 54.14
& 14.20
& 37.70
& 50.70
& 27.70
& 35.60
& 21.80
& 34.40
& 38.19 \\

InternVL-3.5-8B
& 65.09
& 15.49
& 43.06
& 54.52
& 22.99
& 37.02
& 30.13
& 36.41
& 37.43 \\

\midrule


\rowcolor{gray!5}
\multicolumn{10}{c}{\textit{2D Medical MLLMs}} \\

\midrule

Lingshu-7B
& 78.70
& 14.20
& 14.60
& 58.48
& 29.46
& 37.17
& 35.43
& 39.59
& 40.16 \\

HuluMed-7B
& \ftleft{81.60}
& \ftmid{58.00}
& \ftmid{79.40}
& \plainleft{60.06}
& 25.21
& 32.21
& 32.28
& 36.44
& 42.96 \\

MedGemma-1.5-4B
& 66.68
& 17.47
& 19.32
& 36.78
& 26.05
& 28.85
& 31.13
& 30.32
& 38.24 \\

OmniCT-7B
& \ftleft{67.92}
& \ftmid{47.09}
& \ftmid{69.23}
& \plainleft{50.62}
& 24.72
& 30.60
& 27.35
& 32.84
& 36.39 \\

\midrule


\rowcolor{gray!5}
\multicolumn{10}{c}{\textit{3D Medical MLLMs}} \\

\midrule

RadFM-14B-FT
& 29.20
& 44.11
& 42.99
& \ftleft{62.30}
& \ftmid{70.10}
& \ftmid{66.30}
& \ftmid{27.70}
& \ftmid{58.90}
& \plainleft{31.49} \\

M3D-Phi-3-4B-FT
& \ftleft{82.43}
& \ftmid{49.30}
& \ftmid{74.77}
& \ftleft{66.20}
& \ftmid{\textbf{70.30}}
& \ftmid{65.80}
& \ftmid{29.10}
& \ftmid{59.80}
& \plainleft{30.57} \\

Photon-3B-FT
& \ftleft{83.07}
& \ftmid{52.86}
& \ftmid{77.01}
& \plainleft{51.21}
& 19.62
& 32.67
& 27.35
& 32.77
& 33.25 \\

\midrule


\rowcolor{gray!5}
\multicolumn{10}{c}{\textit{Ours}} \\

\midrule

\rowcolor{blue!5}
\textbf{RadSight-4B}
& \ftleft{\textbf{85.14}}
& \ftmid{\underline{63.80}}
& \ftmid{\textbf{81.69}}
& \ftleft{\textbf{85.62}}
& \ftmid{\underline{69.90}}
& \ftmid{\textbf{68.98}}
& \ftmid{\underline{73.54}}
& \ftmid{\textbf{73.40}}
& \plainleft{\textbf{44.33}} \\

\rowcolor{blue!5}
\textbf{RadSight-8B}
& \ftleft{\textbf{85.14}}
& \ftmid{\textbf{63.94}}
& \ftmid{\underline{80.89}}
& \ftleft{\underline{83.74}}
& \ftmid{68.76}
& \ftmid{\underline{68.56}}
& \ftmid{\textbf{74.28}}
& \ftmid{\underline{72.76}}
& \plainleft{\underline{44.06}} \\

\bottomrule

\end{tabular}
}
\vspace{-0.5cm}
\end{table*}

\subsection{Main Results}
\textbf{Performance on Perception-Bench.}
Table~\ref{tab:tumor_bench} presents the results on our Perception-Bench, which evaluates medical MLLMs from low-level perception to high-level clinical outputs. The main observations are as follows. \textbf{(i) Existing medical MLLMs exhibit clear deficiencies in low-level perception.} Although both generalist and medical MLLMs show certain medical image understanding capabilities, their performance remains limited on perception-centric tasks, such as attribute judgment, spatial grounding, and spatial understanding. For example, medical-specific MLLMs such as Lingshu and HuluMed achieve only 38.26 and 40.81 on attribute judgment, while the 3D medical MLLM M3D obtains only 27.36 on spatial grounding. These results indicate that current models still struggle to recognize fine-grained lesion attributes and localize clinically meaningful regions. This supports our hypothesis that the clinical failures of medical MLLMs are closely related to insufficient low-level visual perception. \textbf{(ii) Existing 2D and 3D medical models suffer from modality-specific limitations.} Existing 2D medical MLLMs perform reasonably well on slice-based image understanding but do not natively support volumetric grounding. In contrast, dedicated 3D medical MLLMs are designed primarily for volumetric CT inputs and lack native support for several 2D perception tasks. This fragmented capability limits their applicability to clinical scenarios involving heterogeneous imaging modalities. Through its modality-specific dual-encoder architecture, RadSight supports both 2D images and 3D CT volumes within a unified visual-language interface and achieves strong performance across both modalities. \textbf{(iii) RadSight achieves consistent gains from perception to high-level clinical outputs.} RadSight also substantially improves diagnostic and report-generation performance. RadSight-4B achieves an accuracy of 56.16 on disease prediction and an AUROC of \textbf{81.10} across 82 diseases on anomaly detection, outperforming the strongest baseline by 10.71 points. For report generation, RadSight-4B obtains the highest CE F1 score of 36.45. These results are consistent with our central hypothesis that reliable clinical interpretation benefits from representations that capture fine-grained attributes and spatial evidence learned during the preceding stages. Overall, Perception-Bench reveals the perceptual limitations of existing medical MLLMs, while RadSight substantially narrows this gap through perception-driven training and unified visual modeling.

\textbf{Performance on Public 2D and 3D Benchmarks.}
Tables~\ref{tab:2d_benchmark} and~\ref{tab:3d_benchmark} present the performance of RadSight on public 2D and 3D medical benchmarks, respectively. On the 2D benchmarks, RadSight-8B achieves the best average score of \textbf{83.18}, outperforming the strongest baseline, Lingshu-7B, by \textbf{6.83 points}. RadSight-4B also achieves a competitive average score of \textbf{81.79}, despite its smaller model size. These results establish RadSight as a strong and competitive model across public 2D medical VQA and recognition benchmarks.

RadSight exhibits particularly strong performance on public 3D CT
benchmarks. As shown in Table~\ref{tab:3d_benchmark}, dedicated 3D
medical MLLMs, such as M3D and Photon, achieve competitive results
after benchmark-specific fine-tuning. Several 2D medical MLLMs, such
as HuluMed, can also process CT volumes as sequences of 2D slices.
However, slice-based processing may not fully preserve volumetric
anatomy and inter-slice spatial dependencies, which are important for
comprehensive 3D CT understanding. In contrast, RadSight directly models volumetric CT inputs and outperforms the existing baselines across most evaluation dimensions.
On 3D-Rad, RadSight achieves the best results on all three subtasks,
surpassing both 3D-specialized models and slice-based medical MLLMs.
On DeepTumorVQA, RadSight-4B obtains the highest overall score of
\textbf{73.40}, outperforming the strongest baseline by
\textbf{13.60 points} and demonstrating strong volumetric visual and
clinical reasoning capabilities. Notably, without task-specific fine-tuning on CT-Spatial-VQA, RadSight-4B achieves the best accuracy of 44.33, outperforming the strongest baseline by 1.37 points. These results demonstrate the strong performance of RadSight on public 3D CT benchmarks and its ability to generalize to an unseen task
format without task-specific fine-tuning.
\begin{wraptable}[11]{r}{0.44\textwidth} 
    \centering
    \vspace{-3mm}
    \caption{Performance on DeepTumorVQA without task-specific fine-tuning. \textbf{Bold} and \underline{underlined} values indicate the best and second-best results.}
    \resizebox{0.43\textwidth}{!}{%
    \begin{tabular}{lccccc} 
        \toprule
        \textbf{Model} & \textbf{Recog.} & \textbf{Meas.} & \textbf{Vis.R.} & \textbf{Med.R.} & \textbf{Overall} \\
        \midrule
        \rowcolor{gray!5}
        \multicolumn{6}{c}{\textit{Generalist MLLMs}} \\ 
        \midrule
        Qwen3-VL-8B      & 50.70 & 27.70 & 35.60 & 21.80 & 34.40 \\
        InternVL-3.5-8B    & 54.52 & 22.99 & 37.02 & 30.13 & 36.41 \\
        \midrule
        \rowcolor{gray!5}
        \multicolumn{6}{c}{\textit{Medical MLLMs}} \\ 
        \midrule
        Lingshu-7B     & 58.48 & \textbf{29.46} & \underline{37.17} & \underline{35.43} & \underline{39.59} \\
        HuluMed-7B   & \textbf{60.06} & 25.21 & 32.21 & 32.28 & 36.44 \\
        MedGemma-1.5-4B   & 36.78 & 26.05 & 28.85 & 31.13 & 30.32 \\
        OmniCT-7B    & 50.62 & 24.72 & 30.60 & 27.35 & 32.84 \\
        \midrule
        RadFM-14B    & 39.10  & 21.70  & 25.40  & 26.30 & 27.60 \\
        M3D-Phi-3-4B       & 4.60  & 25.00  & 25.20  & 14.20 & 18.90 \\
        Photon-3B & 51.21 & 19.62 & 32.67 & 27.35 & 32.77 \\
        \rowcolor{blue!5}
        \textbf{RadSight-4B}    & \underline{59.32} & \underline{27.83} & \textbf{39.86} & \textbf{58.01} & \textbf{44.82} \\
        \bottomrule
    \end{tabular}
    }  
    \label{tab:deeptumorVQA}
    \vspace{-3mm}
\end{wraptable}

\textbf{Generalization to DeepTumorVQA without Task-Specific Fine-Tuning.}
Although DeepTumorVQA and part of our training corpus share the same
underlying data source, we use no DeepTumorVQA QA pairs or task-specific
fine-tuning and maintain strict patient-level separation. Since its task
formulations differ from our training data, this evaluation directly
examines whether the fine-grained perceptual capabilities learned by
RadSight can generalize to unseen patients and downstream clinical
tasks. As shown in Table~\ref{tab:deeptumorVQA}, RadSight-4B achieves the best overall score of \textbf{44.82}, outperforming the strongest baseline, Lingshu-7B, by \textbf{5.23 points}. More importantly, it achieves the best
visual reasoning score of \textbf{39.86} and medical reasoning score of
\textbf{58.01}. The improvement in visual reasoning is consistent with
our fine-grained perception training, which explicitly supervises
lesion attributes and spatial evidence, while the strong medical
reasoning performance suggests that the diagnosis stage effectively
connects perceptual evidence with disease-level clinical semantics.
Overall, these results demonstrate that the perceptual and diagnostic
capabilities learned by RadSight generalize to unseen patients and
distinct QA formulations without task-specific fine-tuning.

\subsection{Ablation and Case Study}
\textbf{Effect of Progressive Training Curriculum.}
We conduct an ablation study to examine the contribution of different training stages to RadSight. As shown in Table~\ref{tab:ablation_study}, using only the alignment stage provides a basic visual-language interface, but its clinical correctness and external benchmark performance remain limited. Introducing the fine-grained perception stage brings improvements, especially in report generation and external benchmark evaluation. Specifically, CT-Rate-F1 increases from 15.90 to 26.21, indicating that low-level perception provides essential visual evidence for downstream tasks. Further incorporating the diagnosis stage strengthens disease-level understanding and substantially improves diagnostic metrics. For example, CT-Rate-F1 increases from 15.90 to 32.38. Notably, neither
perception-only nor diagnosis-only training improves CT-Spatial-VQA, achieving 41.17 and 37.10, respectively, compared with 41.66 for alignment alone. Since CT-Spatial-VQA covers both fine-grained perception and clinical interpretation, supervision at either level alone is insufficient. Combining both stages increases the score to 44.33. Training only with diagnosis-level supervision may encourage the model to produce disease-level responses while weakening its attention to low-level perceptual details. This observation further highlights the importance and priority of learning low-level perception. Meanwhile, the comparison between the third and fourth settings shows that high-level diagnostic learning benefits from low-level perceptual grounding, but cannot replace it. Overall, the ablation study validates the effectiveness of our progressive perception-driven curriculum in improving both clinical reliability and external generalization. An additional ablation comparing joint and progressive training is provided in Appendix~\ref{sec:appendix-ablation}.

\textbf{Effect of Dual Vision Encoder Design.}
\begin{figure*}[t]
  \centering
  \includegraphics[width=1\linewidth,clip=false, trim=0 8 5 0]{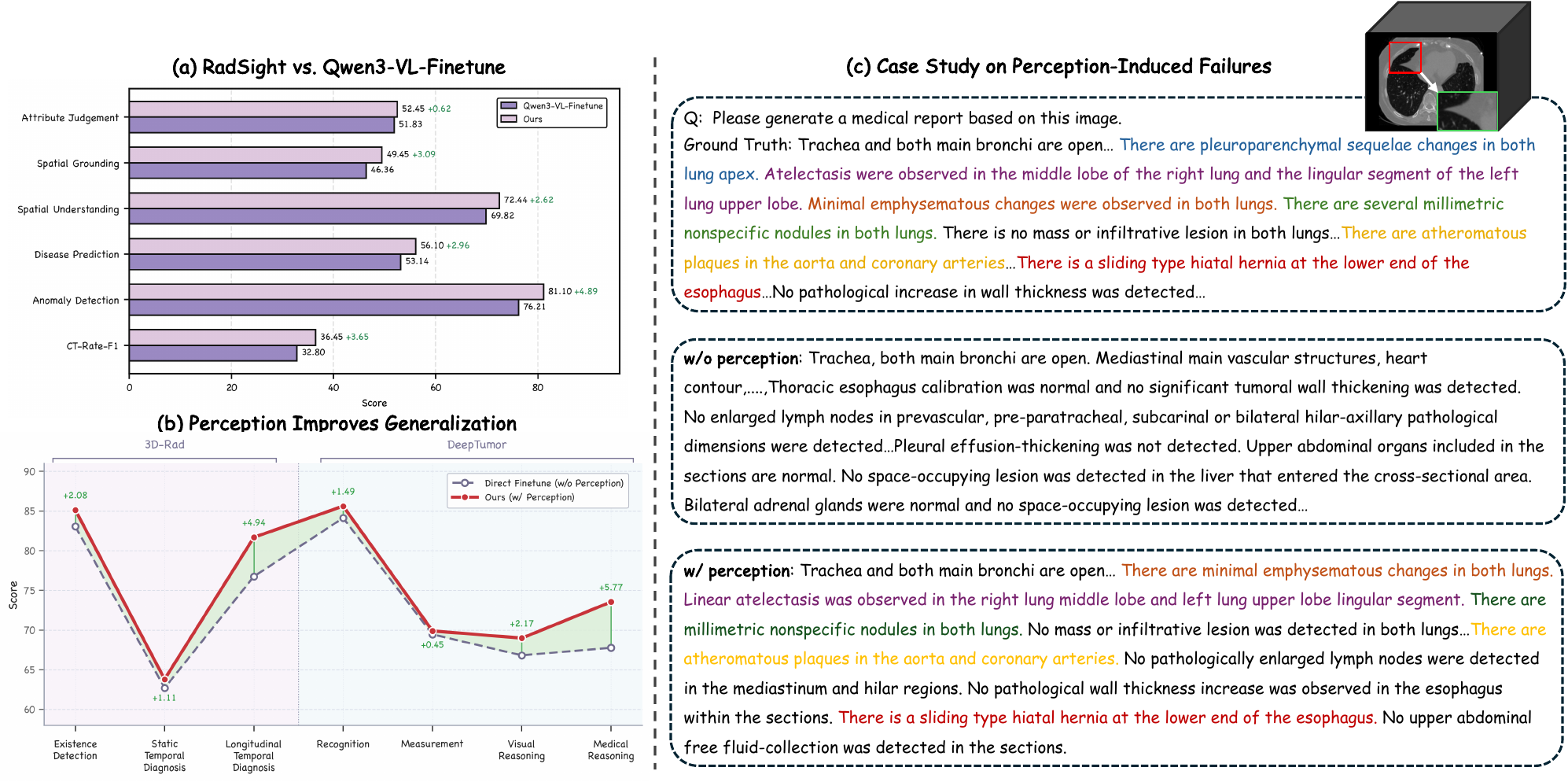}
    \vspace{-2mm}
    \caption{\textbf{Ablation and case study on perception-oriented modeling.}
    (a) Comparison of RadSight and Qwen3-VL-Finetune on perceptual and clinical tasks. (b) Effect of perception-oriented training on public benchmark performance. (c) A qualitative case study showing that insufficient visual perception leads to missed findings, while RadSight recovers clinically relevant abnormalities.
    }
  \label{fig:case_study}
  \vspace{-3mm}
\end{figure*}

To verify whether RadSight’s improvements come from the model design rather than merely from fine-tuning data, we compare it with Qwen3-VL-Finetune (Qwen3-VL-FT), which is directly fine-tuned on the same perception-oriented dataset. As shown in Figure~\ref{fig:case_study}~(a), RadSight consistently outperforms Qwen3-VL-FT across all evaluated dimensions. This suggests that directly adapting a strong generalist MLLM to medical data is still insufficient for reliable medical perception. The advantage is especially clear on tasks requiring spatial and volumetric understanding. In particular, RadSight improves spatial grounding mIoU from 46.35 to 49.45, showing that the dual-encoder design enables more accurate localization of medical regions. These results demonstrate the importance of explicitly modeling 2D and 3D medical inputs with modality-aware visual encoders. By preserving the native spatial structures of slice-based images and volumetric CT scans, RadSight learns more reliable perceptual representations, which further benefit downstream tasks.
\begin{wraptable}[9]{r}{0.60\textwidth} 
    \caption{Ablation analysis on the contribution of different stages. The best performance is highlighted in \textbf{bold}.}
    \label{tab:ablation_study}
    \centering 
    \renewcommand{\arraystretch}{1.0} 
    \resizebox{\linewidth}{!}{
    \begin{tabular}{c c c | c c | c c | c} 
        \toprule
         & & & \multicolumn{2}{c}{\textbf{Diagnosis}} & \multicolumn{2}{c}{\textbf{Report Generation}} & \multicolumn{1}{c}{\textbf{Public Bench}} \\
        \cmidrule(lr){4-5} \cmidrule(lr){6-7} \cmidrule(lr){8-8}
        \textbf{S1} & \textbf{S2} & \textbf{S3} & \textbf{D.P} & \textbf{A.D} & \textbf{RateScore} & \textbf{CT-Rate-F1} & \textbf{CT-SpatialVQA} \\
        \midrule

        \rowcolor{gray!5}
        \checkmark & - & - & -- & -- & 58.45 & 15.90 & 41.66 \\

        \rowcolor{gray!5}
        \checkmark & \checkmark & - & -- & -- & 59.43 & 26.21 & 41.17 \\

        \rowcolor{gray!5}
        \checkmark & - & \checkmark & 54.04 & 77.36 & 60.50 & 32.38 & 37.10 \\

        \rowcolor{blue!5} 
        \textbf{\checkmark} & \textbf{\checkmark} & \textbf{\checkmark} & \textbf{56.16} & \textbf{81.10} & \textbf{60.73} & \textbf{36.45} & \textbf{44.33} \\

        \bottomrule
    \end{tabular}
    }
\end{wraptable}

\textbf{Generalization of Learned Perceptual Representations.}
To evaluate whether the perceptual capabilities learned from our perception-oriented curriculum can be effectively transferred to downstream medical tasks, we further evaluate RadSight on public benchmarks. Notably, these benchmarks adopt task formulations that are substantially different from our training corpus, providing a suitable testbed for assessing the transferability of the learned perceptual representations. As shown in Figure~\ref{fig:case_study}~(b), RadSight consistently outperforms models without perception-oriented training across various 3D CT tasks. The improvements are particularly evident in tasks requiring visual reasoning, such as Longitudinal Temporal Diagnosis in 3D-Rad and both visual reasoning and medical reasoning in DeepTumorVQA. Specifically, RadSight achieves a \textbf{5.77-point improvement} in medical reasoning on DeepTumorVQA, demonstrating that enhanced perceptual representations can be effectively utilized for higher-level clinical interpretation. These results indicate that our perception-oriented curriculum does not merely optimize for specific task formats, but instead enables the model to learn transferable visual representations that benefit diverse downstream CT understanding tasks. 

\textbf{Case Study on Perception-Induced Failures.}
Figure~\ref{fig:case_study}~(c) further illustrates how low-level perception affects high-level clinical outputs. Without perception-oriented training, the model produces a generic report and misses multiple clinically relevant findings. In contrast, RadSight captures these fine-grained abnormalities and generates a report that better matches the ground truth, such as emphysematous changes, atelectasis, small pulmonary nodules, atheromatous plaques, and hiatal hernia. This case highlights the importance of learning low-level lesion attributes and spatial evidence, as such perceptual cues directly influence the completeness and reliability of downstream clinical outputs. Additional case studies on low-level perception and perception-induced failures are provided in the Appendix.
\section{Discussion and Conclusion}
In this work, we investigated whether unreliable clinical outputs from medical multimodal large language models are associated with weaknesses in fine-grained visual perception. We introduced Perception-Bench to evaluate attribute judgment, spatial grounding, spatial understanding, disease prediction, anomaly detection, and radiology report generation. The benchmark revealed persistent perceptual limitations in existing MLLMs. We therefore developed RadSight, which combines dedicated 2D and 3D visual encoders with a progressive perception-driven curriculum. Experiments on Perception-Bench and public medical benchmarks showed consistent improvements over the evaluated baselines. Ablation results further indicated that fine-grained perception and clinical diagnosis supervision are complementary.

These findings identify low-level visual perception as an important contributor to reliable clinical understanding, rather than its sole determinant. The current evaluation is limited to retrospective public radiology datasets and automated metrics, while small-lesion grounding remains challenging. Future work should examine additional imaging modalities, external clinical data, and radiologist-centred evaluation. Within these boundaries, RadSight demonstrates the value of explicitly grounding medical multimodal understanding in fine-grained visual evidence.

\section*{Acknowledgement}
This work was supported by the Zhejiang Province Postdoctoral Research Excellence Funding Program (ZJ2024032).

\bibliographystyle{assets/plainnat}
\bibliography{paper}

\newpage
\beginappendix
\section*{Appendix}
\label{sec:appendix}
\Cref{sec:appendix-LLM}. LLM Usage Statement.

\cref{sec:appendix-data}. Data Construction Details. 

\quad \Cref{sec:appendix-data_sources}. Data Sources.

\quad \Cref{sec:appendix-data_split}. Patient-Level Dataset Splits and Leakage Prevention.

\quad \Cref{sec:appendix-report_parsing}. Prompt Design of Perception-oriented QA Generation.


\quad \Cref{sec:appendix-data_dist}. Training Data Distribution Details.

\quad \Cref{sec:appendix-perception_bench_dist}. Perception-Bench Distribution.

\Cref{sec:appendix-implementation}. Experiments and Implementation Details.

\quad \Cref{sec:appendix-data_process}. Data Processing Details.

\quad \Cref{sec:appendix-model}. Model Architecture Details.

\quad \Cref{sec:appendix-public_bench}. Public Benchmarks Details.

\quad \Cref{sec:appendix-training_details}. Training Details.

\Cref{sec:appendix-metrics}. Evaluation Metrics.

\quad \Cref{sec:appendix-attribute_eval_prompt}. Prompt-based LLM Evaluation for Attribute Judgment.

\quad \Cref{sec:appendix-grounding_metrics}. Grounding Metrics.

\quad \Cref{sec:appendix-clinical_metrics}. Clinical Evaluation Metrics.

\quad \Cref{sec:appendix-mrg_metrics}. Report Generation Evaluation Metrics.

\Cref{sec:appendix-results}. Additional Experimental Results.

\quad \Cref{sec:appendix-ablation}. Ablation Studies.

\quad \Cref{sec:appendix-result_details}. Full Experimental Results.

\Cref{sec:appendix-cases}. Cases Study.

\newpage

\section{LLM Usage Statement}
\label{sec:appendix-LLM}
LLMs are used in this work only for controlled assistance in data construction, quality refinement, and manuscript proofreading. During dataset construction, LLMs assist in extracting and organizing clinical information from radiology reports to facilitate the creation of perception-oriented supervision. LLMs are also used to improve the consistency of generated annotations and refine language expression in the manuscript. All model architectures, experiments, evaluations, and conclusions are designed and verified by the authors.

\section{Data Construction Details}
\label{sec:appendix-data}

\subsection{Data Sources}
\label{sec:appendix-data_sources}

Table~\ref{tab:data_sources} summarizes the datasets used in this work, spanning both 2D chest X-ray and 3D CT
modalities. We categorize each dataset by its annotation type: \textit{Image-Report} denotes datasets with paired radiology
reports, which we leverage for perception-oriented QA generation via our two-stage prompting pipeline, and \textit{Image-Mask} denotes datasets with pixel/voxel-level segmentation annotations, which we use for grounding data construction.

For the 2D X-ray modality, MIMIC-CXR~\citep{mimic} and IU-Xray~\citep{iu_xray} serve as our primary report-based
sources, covering a diverse range of thoracic pathologies across over 235,000 studies. PadChest-GR~\citep{de2025padchest_gr}, VinDr-CXR~\citep{nguyen2022vindr}, MS-CXR~\citep{boecking2022ms-cxr}, and REFLACX~\citep{bigolin2022reflacx} provide pixel-level annotations for anatomical structures and lesion regions, enabling spatial grounding supervision.

For the 3D CT modality, Merlin~\citep{blankemeier2024merlin}, CT-RATE~\citep{ct-rate}, Inspect~\citep{inspect}, and
AMOS-MM~\citep{amos} provide paired radiology reports covering chest and abdominal findings, from which we extract structured disease annotations and attribute-level QA pairs. AbdomenAtlas 3.0~\citep{radgpt} contributes voxel-level segmentation masks for abdominal organs and lesions. The National Lung Screening Trial (NLST)~\citep{nlst} is used for pretraining, providing a large-scale collection of over 73,000 low-dose chest CT studies.

In total, our data curation pipeline covers approximately 150,000 patients and over 440,000 studies, ensuring broad anatomical coverage and pathological diversity across both modalities.

\begin{table*}[t]
\centering
\caption{Summary of datasets used in this work. ``-'' indicates that the corresponding statistic is not applicable or
not reported by the original dataset.}
\label{tab:data_sources}
\adjustbox{max width=\textwidth}{%
\begin{tabular}{@{}llccll@{}}
\toprule
\textbf{Modality} & \textbf{Dataset} & \textbf{\#Patients} & \textbf{\#Studies} & \textbf{Type} & \textbf{URL} \\
\midrule
\multirow{5}{*}{\textit{2D (X-ray)}}
& MIMIC-CXR~\citep{mimic}       & 65,379  & 227,835 & Image-Report &
\url{https://physionet.org/content/mimic-cxr/}         \\
& IU-Xray~\citep{iu_xray}       & --   & 7,470   & Image-Report &
\url{https://www.kaggle.com/datasets/raddar/chest-xrays-indiana-university} \\
& PadChest-GR~\citep{de2025padchest_gr}       & --  & 4,555 & Image-Mask &
\url{https://bimcv.cipf.es/bimcv-projects/padchest-gr/}   \\
& VinDr-CXR~\cite{nguyen2022vindr}        & --      & 15,000  & Image-Mask   & \url{https://vindr.ai/datasets/cxr} \\
& MS-CXR~\citep{boecking2022ms-cxr}        & --      & 1026   & Image-Mask &
\url{https://physionet.org/content/ms-cxr/1.1.0/}            \\
& REFLACX~\citep{bigolin2022reflacx}        & --      & 3,056   & Image-Mask &
\url{https://physionet.org/content/reflacx/}            \\
\midrule
\multirow{6}{*}{\textit{3D (CT)}}
& Merlin~\citep{blankemeier2024merlin}      & 18,317    & 25,494    & Image-Report &
\url{https://stanfordaimi.azurewebsites.net/datasets/60b9c7ff-877b-48ce-96c3-0194c8205c40}
              \\
& CT-RATE~\citep{ct-rate}     & 21,304  & 50,188  & Image-Report &
\url{https://huggingface.co/datasets/ibrahimhamamci/CT-RATE} \\
& Inspect~\citep{inspect}                      & 19,402    & 23,248    & Image-Report         &
\url{https://stanfordaimi.azurewebsites.net/datasets?term=INSPECT} \\
& AbdomenAtlas 3.0~\citep{radgpt} & --    & 9,262   & Image-Report   &
\url{https://huggingface.co/datasets/AbdomenAtlas/AbdomenAtlas3.0Mini} \\
& AMOS-MM~\citep{amos}               & --      & 1,690     & Image-Report   &
\url{https://era-ai-biomed.github.io/amos/dataset.html#download} \\
& National Lung Screening Trial(NLST)~\citep{nlst} & 26,722    & 73,116   & --   &
\url{https://www.cancerimagingarchive.net/wp-content/uploads/manifest-NLST_allCT.tcia} \\
\bottomrule
\end{tabular}%
}
\end{table*}

\subsection{Patient-Level Dataset Splits and Leakage Prevention}
\label{sec:appendix-data_split}

Except for NLST~\citep{nlst}, we followed the official splits provided by each source dataset. All datasets were partitioned at the patient level, ensuring that all studies from the same patient were assigned exclusively to one subset. For NLST, we created patient-level training and held-out subsets using a 9:1 ratio.

Notably, both our training corpus and DeepTumorVQA~\citep{chen2026deeptumorvqa} include data from AbdomenAtlas~3.0~\citep{radgpt}. For this shared source, we adopted exactly the same patient-level partition as DeepTumorVQA. Consequently, no patient assigned to the DeepTumorVQA evaluation split was included in our training data, preventing patient-level data leakage between training and evaluation.

\subsection{Prompt Design of Perception-oriented QA Generation}
\label{sec:appendix-report_parsing}
We employ a two-stage prompting pipeline to convert free-text radiology reports into structured disease-level annotations with fine-grained attributes. In the first stage, the large language model (LLM) is instructed to perform organ-level decomposition followed by disease identification. Specifically, the model first segments the original
report into organ-level sub-reports by mapping all descriptive sentences to standardized anatomical entities (e.g., "lung," "liver," "heart"), resolving anaphora and duplicating cross-organ sentences as needed to ensure each sub-report is self-contained. The model then identifies all diseases or clinical findings within each organ sub-report, normalizing synonymous or laterality-specific terms into standardized disease names (e.g., "left lower lobe opacity"
and "bilateral opacities" are unified as "Lung Opacity"). Each identified finding is recorded with its associated organ, supporting evidence sentences, and a status label from {present, absent, suspected, historical}. In the second stage, conditioned on each extracted disease and a predefined set of candidate attributes, the model performs attribute extraction and question–answer (QA) pair generation. The model first determines which candidate attributes are
explicitly described in the report for the target disease, then extracts the corresponding values, and finally constructs natural-language QA pairs grounded solely in the report text. Strict constraints are imposed throughout both stages to prevent hallucination: the model is required to use only information explicitly stated in the report, avoid merging or inferring attributes, and ensure each extracted value is directly supported by the source text. The diseases extracted from each dataset are summarized in Table~\ref{tab:disease_list}. The complete prompt templates for both stages are provided in Figure~\ref{fig:disease-extract} and Figure~\ref{fig:qa-make}, respectively.

\begin{table*}[t]
\centering                    
\caption{Summary of diseases extracted from each dataset.}
\label{tab:disease_list}
\resizebox{\textwidth}{!}{%
\begin{tabular}{@{}llp{13cm}c@{}}
\toprule
\textbf{Modality} & \textbf{Dataset} & \textbf{Diseases} & \textbf{\#} \\
\midrule
\multirow{5}{*}{\textit{2D}}
& MIMIC-CXR
& Atelectasis, Atherosclerosis, Cardiomegaly, Consolidation, Edema, Emphysema, Enlarged Cardiomediastinum, Fracture,
Hernia, Lung Lesion, Lung Opacity, Pleural Effusion, Pneumonia, Pneumothorax, Pulmonary Edema, Support Devices
& 16 \\
\cmidrule(l){2-4}
& IU-Xray
& Atelectasis, Atherosclerosis, Cardiomegaly, Degenerative Changes, Emphysema, Fracture, Lung Opacity, Pleural
Effusion, Scoliosis, Support Devices
& 10 \\
\midrule
\multirow{13}{*}{\textit{3D}}
& CT-RATE
& Aortic Aneurysm, Arterial Wall Calcification, Atelectasis, Bronchiectasis, Cardiomegaly, Consolidation, Coronary
Artery Disease, Coronary Calcification, Emphysema, Fibrosis, Gallstones, Hepatic Steatosis, Hepatomegaly, Hernia,
Interlobular Septal Thickening, Kidney Stone, Kyphosis, Liver Lesion, Lung Nodule, Lung Opacity, Lymphadenopathy,
Medical Material, Mosaic Attenuation Pattern, Osteophyte, Peribronchial Thickening, Pericardial Effusion, Pleural
Effusion, Pneumonia, Pulmonary Fibrotic Sequela, Renal Cyst, Scoliosis, Splenomegaly, Thyroid Nodule
& 33 \\
\cmidrule(l){2-4}
& Merlin
& Abscess, Anasarca, Aortic Valve Calcification, Appendicitis, Ascites, Atelectasis, Atherosclerosis, Biliary Ductal
Dilation, Bladder Wall Thickening, Bowel Obstruction, Bronchiectasis, Cardiomegaly, Cirrhosis, Colitis, Coronary
Calcification, Degenerative Disc Disease, Diverticulitis, Diverticulosis, Emphysema, Fracture, Free Air, Gallbladder
Wall Thickening, Gallstones, Granuloma, Hemangioma, Hematoma, Hepatic Steatosis, Hepatomegaly, Hernia, Hydronephrosis,
Lung Nodule, Lung Opacity, Lymphadenopathy, Osteopenia, Ovarian Cyst, Pancreatic Atrophy, Pancreatic Cyst,
Pericardial Effusion, Pleural Effusion, Post-operative, Prostatomegaly, Renal Atrophy, Renal Calculus, Renal Cyst,
Renal Hypodensities, Scoliosis, Splenomegaly, Submucosal Edema, Surgically Absent Gallbladder, Thrombosis, Uterine
Fibroids, Varices
& 52 \\
\cmidrule(l){2-4}
& Inspect
& Ascites, Atelectasis, Bronchiectasis, Cardiomegaly, Emphysema, Fracture, Hepatic Steatosis, Hernia, Lung Nodule,
Lung Opacity, Lymphadenopathy, Pericardial Effusion, Pleural Effusion, Pneumonia, Pneumothorax, Pulmonary Edema,
Pulmonary Embolism, Pulmonary Hypertension, Splenomegaly, Thyroid Nodule
& 20 \\
\cmidrule(l){2-4}
& AbdomenAtlas
& Colon Lesion, Hepatomegaly, Kidney Enlargement, Kidney Lesion, Liver Lesion, Pancreas Lesion, Pancreatomegaly,
Splenomegaly
& 8 \\
\cmidrule(l){2-4}
& AMOS-MM
& Ascites, Gallstones, Hepatic Steatosis, Hydronephrosis, Lung Opacity, Lymphadenopathy, Prostatomegaly, Renal Cyst,
Renal Hypodensities
& 9 \\
\bottomrule
\end{tabular}%
}
\end{table*}

\begin{figure*}[t]
\begin{tcolorbox}[
    colback=black!5,  
    colframe=black!75, 
    title=Prompt Design for Organ-Level Disease Extraction, 
    fonttitle=\bfseries,
    arc=2mm, 
]
\noindent 

You are a medical report analysis assistant. I will provide you with a CT/X-Ray imaging report, and you are to process it according to the following steps. Do not provide any summaries or explanations. \\

\textbf{Step 1:} Decompose the report into organ-level sub-reports according to the following requirements:

\begin{enumerate}
    \item Carefully read the entire CT report and identify all organs or anatomical structures mentioned.
    
    \item Use only high-level, standardized organ names (e.g., ``lung'', ``liver'', ``kidney'', ``heart'', ``spine''). Do NOT use laterality-specific or region-specific terms as separate organs. For example, ``left lower lobe'', ``right upper lobe'' should all be grouped under ``lung''; ``right arm'', ``left leg'' should be grouped under their parent structure (e.g., ``bone'' or ``soft tissue'').

    \item For each identified organ, extract all sentences or clauses that describe it, forming a dedicated sub-report for that organ.

    \item Ensure that each organ-level sub-report is self-contained and complete, including all relevant descriptions such as location, size, shape, density, and any abnormalities mentioned for that organ.

    \item If a sentence involves multiple organs, duplicate it into each relevant organ's sub-report.

    \item Do not use pronouns; perform anaphora resolution to ensure each sentence is explicit and clear.

    \item Output the result as a JSON object where each key is the standardized organ name and the value is a list of sentences describing that organ. Use \texttt{<step1>} and \texttt{</step1>} as tags.
\end{enumerate}

\textbf{Step 2:} Based on the organ-level sub-reports from Step 1, identify all diseases. For each organ, adhering to the following requirements:

\begin{enumerate}
    \item Use only high-level, standardized disease names (e.g., Pleural Effusion, Lung Opacity, Cardiomegaly, Pneumonia, Lung nodule). Merge synonymous or overly specific findings into a single standard term. For example: ``left lower lobe opacity'', ``right lung opacity'', ``bilateral opacities'' $\rightarrow$ ``Lung Opacity''. ``small pleural effusion'', ``bilateral pleural effusion'' $\rightarrow$ ``Pleural Effusion''. ``PICC line'', ``A port catheter extending to the right atrium'', ``central venous catheter'' $\rightarrow$ ``Medical material''. Do NOT create separate disease entries for left/right side or mild/severe variations.

    \item For each identified disease or finding, record the following fields:
    
    \begin{center}
    \renewcommand{\arraystretch}{1.3}
    \begin{tabular}{>{\ttfamily}l p{0.65\textwidth}}
    \toprule
    \textbf{Field} & \textbf{Description} \\
    \midrule
    organ    & The standardized organ name (from Step 1) \\
    disease  & The standardized disease or finding name \\
    evidence & A list of sentences from Step 1 that support this diagnosis \\
    status   & One of: \texttt{present}, \texttt{absent}, \texttt{suspected}, or \texttt{historical} \\
    \bottomrule
    \end{tabular}
    \end{center}

    \item If an organ shows no abnormalities, record it with an empty disease list.

    \item Output as a JSON list of organ-disease objects. Use \texttt{<step2>} and \texttt{</step2>} as tags.
\end{enumerate}

\end{tcolorbox}
\caption{Prompt Design for Organ-Level Disease Extraction.}
\label{fig:disease-extract}
\end{figure*}

\begin{figure*}[t]
\begin{tcolorbox}[
    colback=black!5,  
    colframe=black!75, 
    title=Prompt Template for Disease Attribute Extraction and QA Generation, 
    fonttitle=\bfseries,
    arc=2mm, 
]
\noindent
CT Report: \texttt{<report>} \\
Disease/Finding: \texttt{<disease>} \\
Candidate Attributes: \texttt{<candidate\_features>} \\
The CT report is already known to mention the disease/finding \texttt{<disease>}. Your task is to identify which attributes of this disease/finding are explicitly described in the report, extract their values, and then construct question-answer pairs based on the extracted information. \\

Please follow the 3 steps below and output each step on a new line only. Do not output any explanation, summary, or extra text. \\

\textbf{General requirements:}
\begin{enumerate}
    \item Only use information explicitly stated in the CT report.
    \item Do not infer, assume, or hallucinate any attribute or value.
    \item Only select attributes from the provided Candidate Attributes.
    \item If multiple findings are described, only extract attributes that correspond to the target disease/finding \texttt{<disease>}.
    \item Keep attribute values concise and accurate.
    \item Do not merge multiple attributes into one.
    \item Each extracted attribute should correspond to one best value only.
\end{enumerate}

\textbf{Step 1:} Identify which candidate attributes of \texttt{<disease>} are explicitly described in the report. \\
If none are described, output exactly: \texttt{<step1>[]</step1>} \\
Otherwise, output exactly a JSON list of attribute names, for example: \\
\texttt{<step1>["Location", "Severity", "Morphology"]</step1>} \\

\textbf{Step 2:} Extract the corresponding value for each attribute identified in Step 1. \\
If Step 1 is empty, output exactly: \texttt{<step2>[]</step2>} \\
Otherwise, output exactly a JSON list of objects in this format: \\
\texttt{<step2>[\{"feature": "Location", "value": "left lower lobe"\}, \{"feature": "Severity", "value": "mild"\}]</step2>} \\
Each feature should appear only once. \\
The value must be directly supported by the report. \\

\textbf{Step 3:} For each feature-value pair in Step 2, construct one question-answer pair. \\
If Step 2 is empty, output exactly: \texttt{<step3>[]</step3>} \\
Otherwise, output exactly a JSON list of objects in this format: \\
\texttt{<step3>[\{"feature": "Location", "question": "Where is the atelectasis located?", "answer": "left lower lobe"\}]</step3>} \\

\textbf{Requirements for Step 3:}
\begin{enumerate}
    \item Each question must ask about one specific attribute of the target disease/finding.
    \item The question must not contain the answer or its paraphrase.
    \item The answer must exactly match the value extracted in Step 2.
    \item Questions should be natural and medically meaningful.
\end{enumerate}

\end{tcolorbox}
\caption{Prompt template for disease attribute extraction and QA generation.}
\label{fig:qa-make}
\end{figure*}


\subsection{Training Data Distribution Details}
\label{sec:appendix-data_dist}
\begin{figure*}[t]
\centering
\includegraphics[width=1\linewidth, clip=false, trim=0 8 5 0]{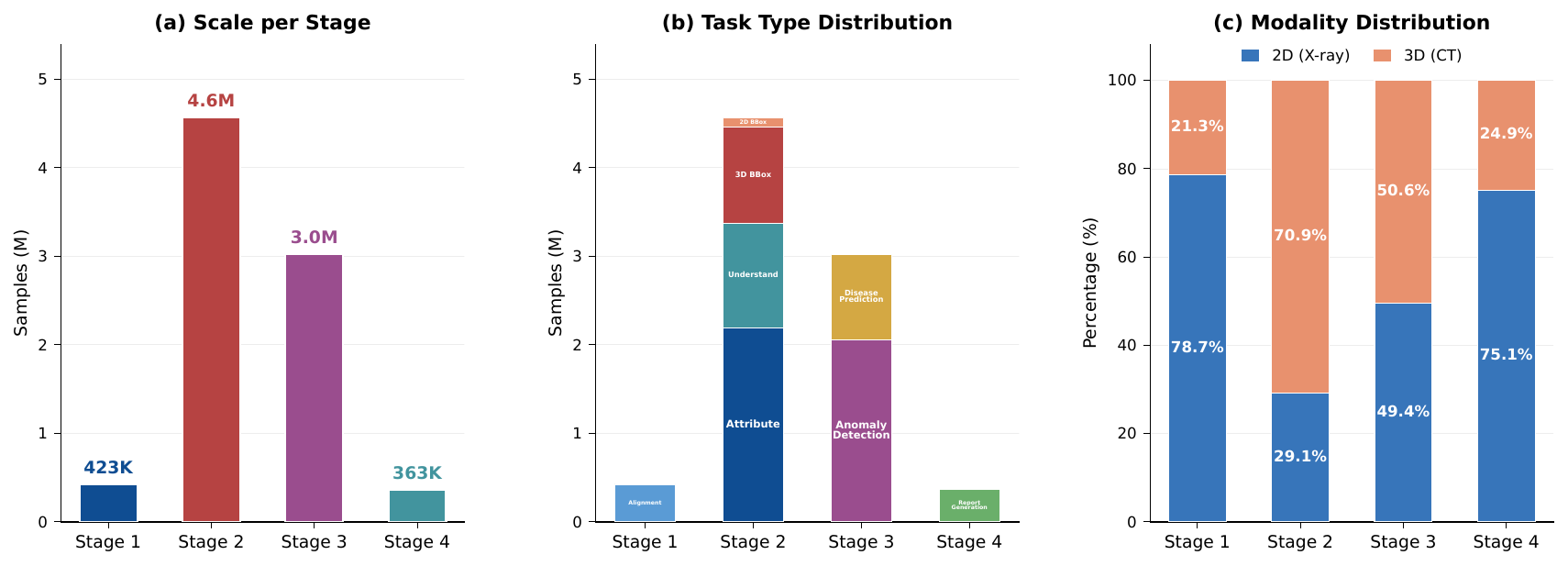}
\caption{Overview of training data across four stages. 	\textbf{(a)} Total sample count per stage. \textbf{(b)} Task type distribution: Stage~1 focuses on vision-language alignment, Stage~2 on perception tasks (attribute QA, understanding, and 2D/3D bounding box localization), Stage~3 on clinical diagnosis (anomaly detection and disease prediction), and Stage~4 on medical report generation. \textbf{(c)} Modality distribution (2D X-ray vs.\ 3D CT) at each stage.}
\label{fig:stage_overview}
\end{figure*}

\begin{figure*}[t]
\centering                    
\includegraphics[width=1\linewidth, clip=false, trim=0 8 5 0]{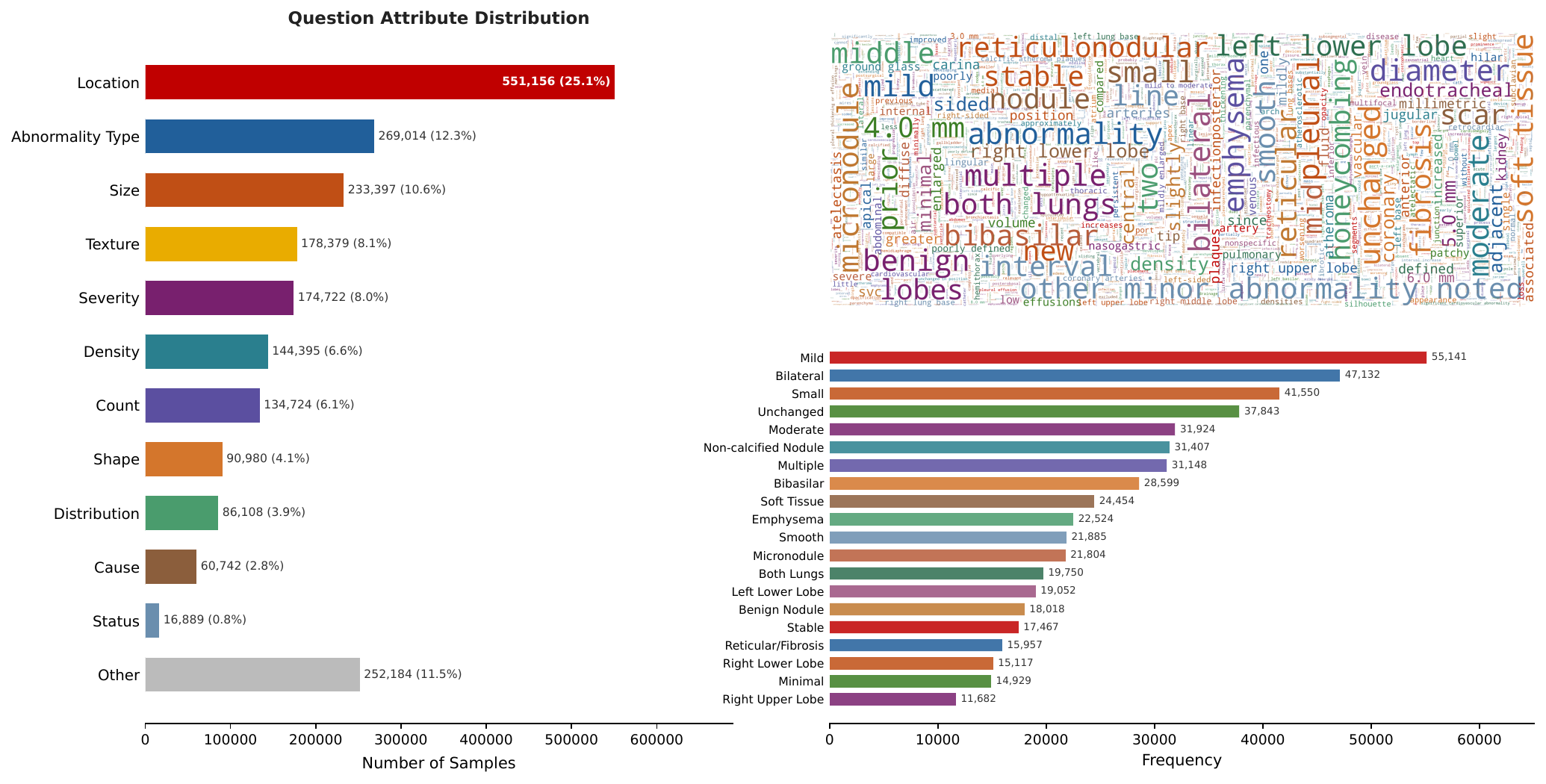}
\caption{Stage~2 attribute judgement VQA data analysis. \textbf{Left:} Distribution of question attributes across 12 categories, with Location (25.1\%) being the most frequent. \textbf{Right top:} Word cloud of medical entities extracted from QA pairs. \textbf{Right bottom:} Top-20 most frequent medical entities and their occurrence counts.}
\label{fig:stage2_attr}
\end{figure*}

\begin{figure*}[t]
\centering
\includegraphics[width=1\linewidth, clip=false, trim=0 8 5 0]{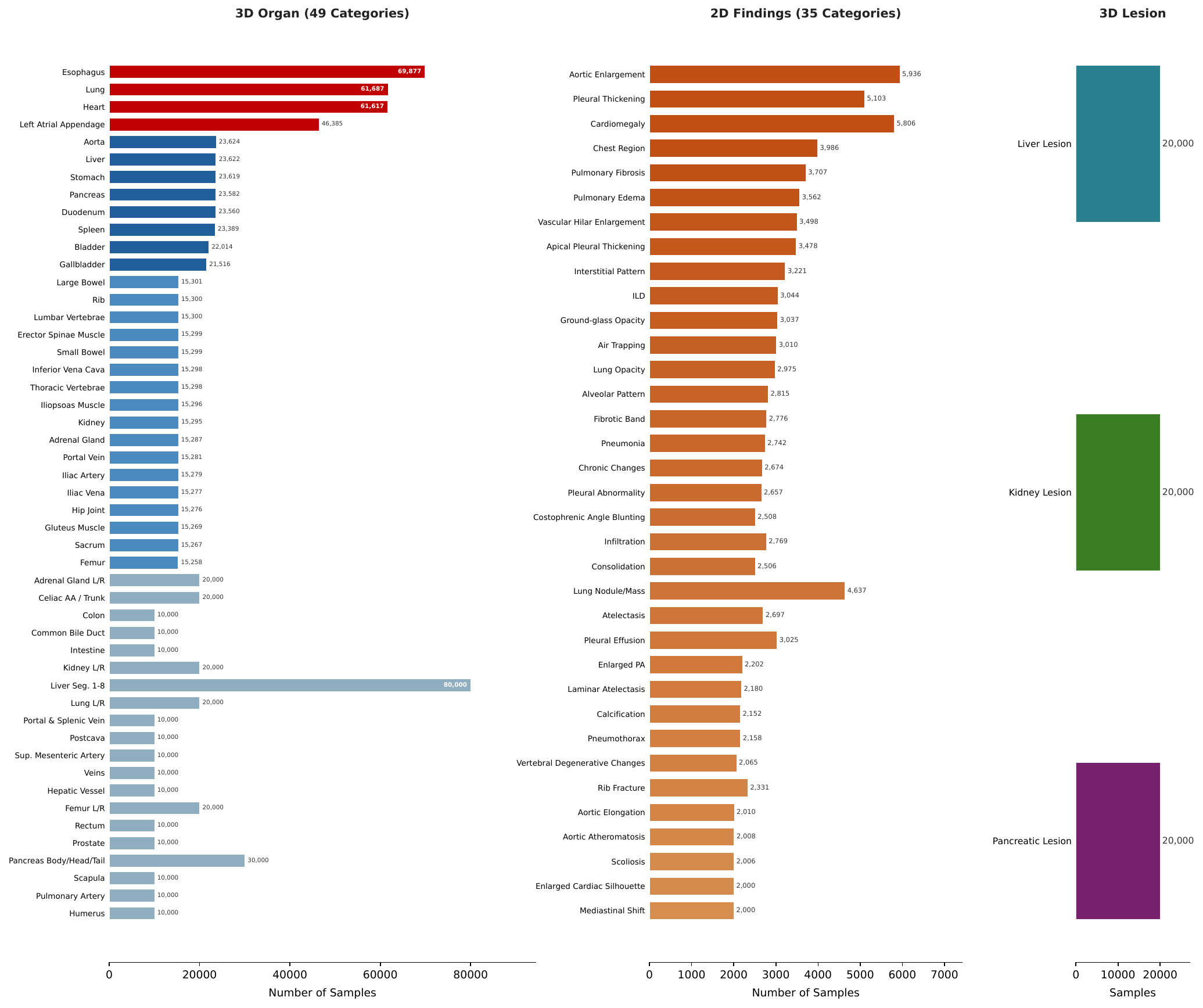}
\caption{Distribution of Stage~2 spatial grounding training data. \textbf{Left:} 3D organ categories (49 types) extracted from CT volumes, covering major anatomical structures across the thorax and abdomen. \textbf{Middle:} 2D radiographic findings (35 types) from chest X-rays, including common abnormalities such as cardiomegaly, pleural effusion, and consolidation. \textbf{Right:} 3D lesion categories (3 types) with 20K samples each for liver, kidney, and pancreatic lesions.}
\label{fig:stage2_organ_lesion}
\end{figure*}

\begin{figure*}[t]
\centering
\includegraphics[width=1\linewidth, clip=false, trim=0 8 5 0]{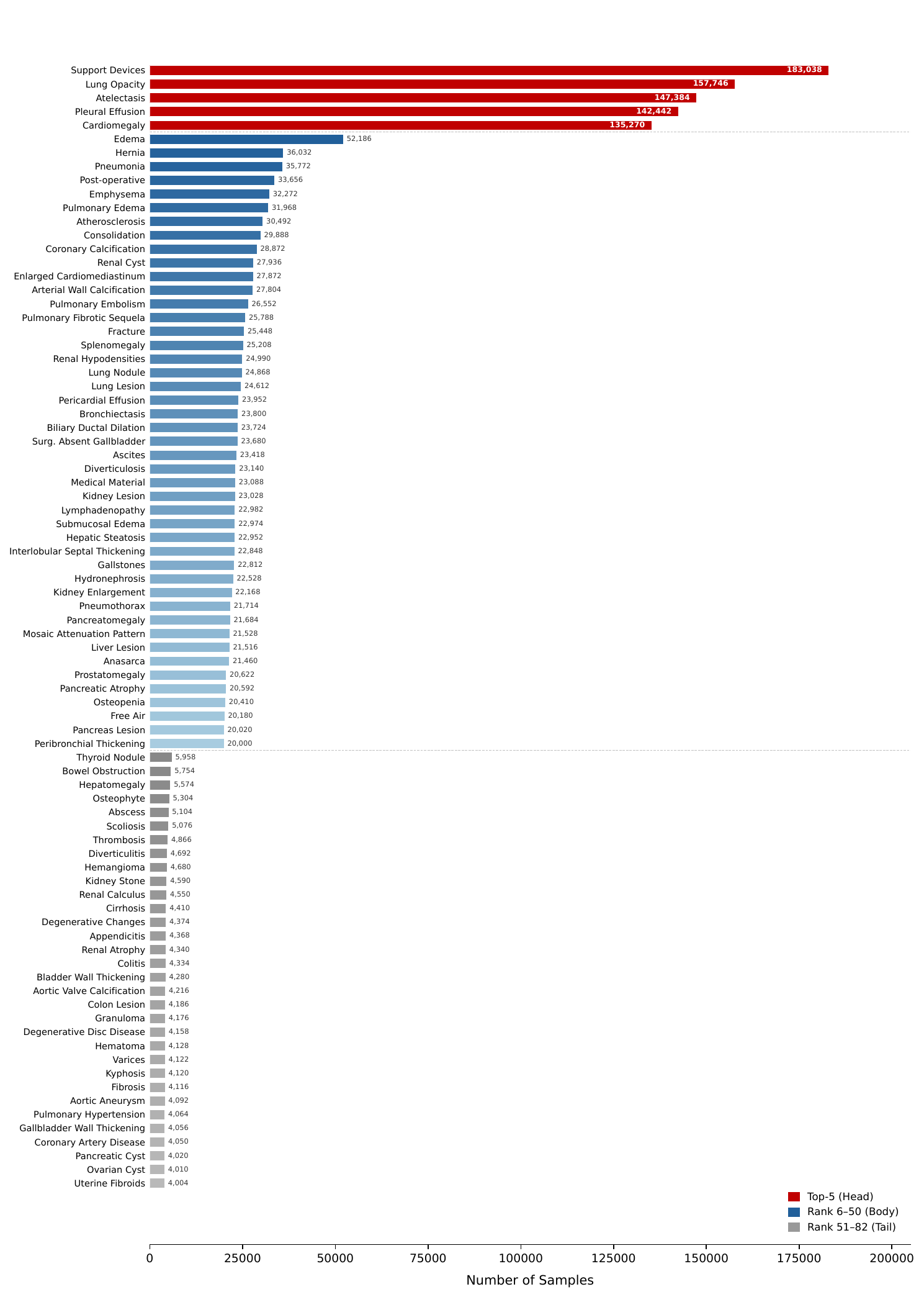}
\caption{Disease distribution of the Stage~3 anomaly detection training data across 82 categories. The top-5 categories (red) are predominantly chest X-ray findings and account for over 37\% of all samples. The body (blue, Rank 6--50) covers a broad spectrum of thoracic and abdominal pathologies, while the tail (gray, Rank 51--82) comprises 32 less common categories.}
\label{fig:stage3_disease}
\end{figure*}

We provide a detailed breakdown of the training data distribution across all four stages in Figures~\ref{fig:stage_overview}--\ref{fig:stage3_disease}. As shown in Figure~\ref{fig:stage_overview}, our training pipeline follows a curriculum learning strategy with progressively specialized objectives. Stage~1 (423K samples) performs vision-language alignment on radiology image-report pairs, with 78.7\% from 2D X-rays and 21.3\% from 3D CT volumes. Stage~2 (4.6M samples) is the largest stage, focusing on fine-grained perception tasks including attribute-level VQA, spatial grounding, and visual understanding, with the modality ratio shifting toward 3D (70.9\%). Stage~3 (3.0M samples) targets clinical diagnosis through anomaly detection and disease prediction, with a roughly balanced 2D/3D split (49.4\% vs.\ 50.6\%). Stage~4 (363K samples) trains on medical report generation from both chest X-rays (75.1\%) and CT volumes (24.9\%). Figure~\ref{fig:stage2_attr} presents the distribution of Stage~2 attribute VQA data. Among the 12 question categories, Location (25.1\%) is the most frequent, followed by Abnormality Type (12.3\%), Size (10.6\%), Texture (8.1\%), and Severity (8.0\%), reflecting the clinical priorities in radiology interpretation. Figure~\ref{fig:stage2_organ_lesion} illustrates the spatial grounding data, which spans 49 3D organ categories, 35 2D radiographic findings, and 3 lesion types (liver, kidney, and pancreatic lesions with 20K samples each). The organ categories cover major anatomical structures from the thorax to the pelvis, providing comprehensive spatial supervision for 3D localization. Figure~\ref{fig:stage3_disease} shows the disease distribution of Stage~3 anomaly detection data across 82 categories. 

\subsection{Perception-Bench Distribution}
\label{sec:appendix-perception_bench_dist}

\begin{figure*}[t]
    \centering
    \includegraphics[width=1\linewidth]{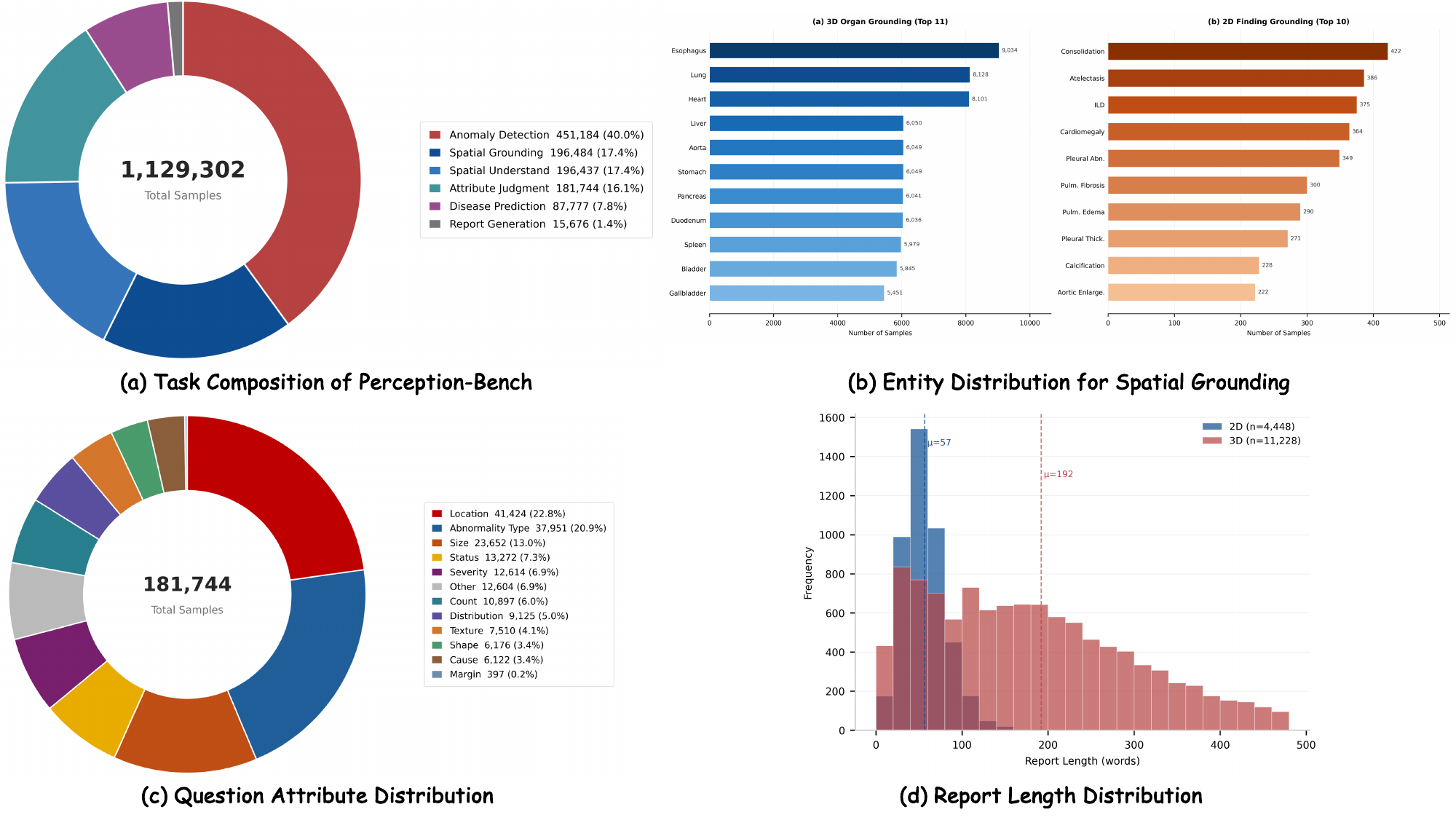}
    \caption{
    \textbf{Composition of Perception-Bench.}
    \textbf{(a)} Distribution of the 1,129,302 evaluation instances across six tasks.
    \textbf{(b)} Most frequent 3D anatomical structures and 2D radiographic findings in the spatial-grounding subset.
    \textbf{(c)} Distribution of the 181,744 attribute-judgment instances across 12 attribute types.
    \textbf{(d)} Report-length distributions for the 2D and 3D report-generation subsets; dashed lines indicate the corresponding mean lengths.
    }
    \label{fig:perception_bench_dist}
\end{figure*}

Perception-Bench contains 1,129,302 evaluation instances across six tasks. As shown in Figure~\ref{fig:perception_bench_dist}(a), anomaly detection is the largest component, with 451,184 instances (40.0\%). Spatial grounding and spatial understanding contain 196,484 (17.4\%) and 196,437 (17.4\%) instances, respectively, followed by attribute judgment with 181,744 (16.1\%), disease prediction with 87,777 (7.8\%), and report generation with 15,676 (1.4\%) instances.

The remaining panels characterize the internal composition of the perception and report-generation tasks. Figure~\ref{fig:perception_bench_dist}(b) reports the most frequent 3D anatomical structures and 2D radiographic findings used for spatial grounding. Figure~\ref{fig:perception_bench_dist}(c) shows that the attribute-judgment subset spans 12 attribute types, with location (22.8\%) and abnormality type (20.9\%) being the most frequent. Figure~\ref{fig:perception_bench_dist}(d) summarizes report lengths for 4,448 2D reports and 11,228 3D reports. The 3D reports are longer on average than the 2D reports (192 versus 57 words).

\begin{figure*}[p]
    \centering
    \includegraphics[height=0.82\textheight]{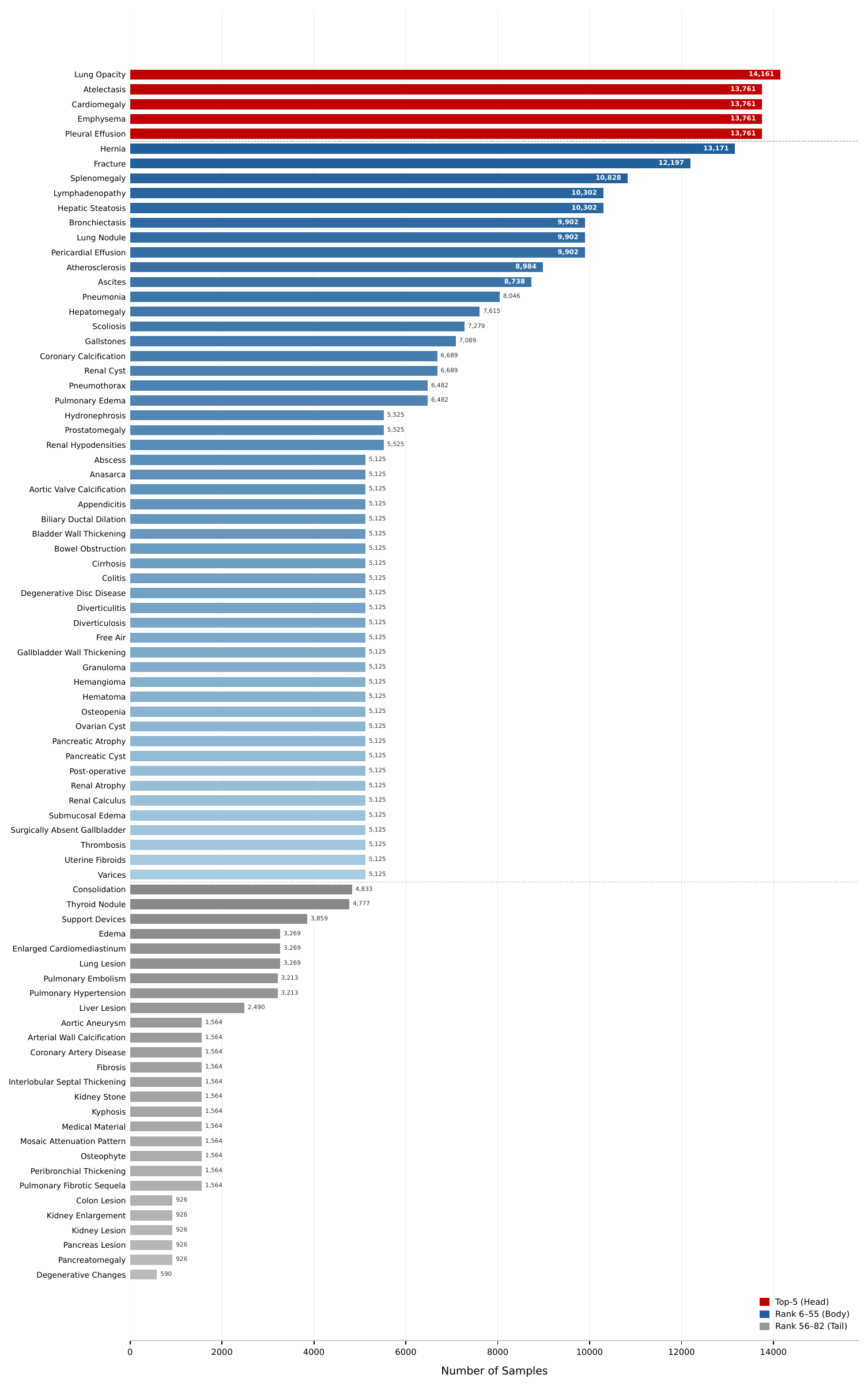}
    \caption{
    \textbf{Category distribution of the Perception-Bench anomaly-detection subset.}
    The subset contains 451,184 instances across 82 abnormality and imaging-finding categories.
    Categories are ordered by sample count and divided into the five most frequent categories (red), body categories ranked 6--55 (blue), and tail categories ranked 56--82 (gray).
    }
    \label{fig:perception_bench_disease_dist}
\end{figure*}

The anomaly-detection subset contains 451,184 evaluation instances covering 82 abnormality and imaging-finding categories. Figure~\ref{fig:perception_bench_disease_dist} shows a long-tailed category distribution. The five most frequent categories contain 69,205 instances (15.3\%), the body categories ranked 6--55 contain 325,799 instances (72.2\%), and the tail categories ranked 56--82 contain 56,180 instances (12.5\%). This distribution documents the breadth and frequency variation of the clinical concepts evaluated by Perception-Bench.

\section{Experiments and Implementation Details}
\label{sec:appendix-implementation}

\subsection{Data Processing Details}
\label{sec:appendix-data_process}
\textbf{3D Volume Preprocessing.}
For 3D CT datasets that do not provide voxel-level annotations, we first run TotalSegmentator~\citep{wasserthal2023totalsegmentator} on each CT volume to generate dense anatomical segmentation masks. The masks and their corresponding CT volumes are resampled to an isotropic in-plane resolution of $1.0 \times 1.0\,\text{mm}^2$ with a slice thickness of $5.0\,\text{mm}$, using trilinear interpolation for images and nearest-neighbor interpolation for masks. CT intensities are then clipped to $[-1000, 1000]$\,HU and linearly normalized to $[0, 1]$. We compute the tight bounding box of all non-zero mask voxels, extend it by 5 voxels along the depth axis and 20 voxels along the height and width axes to preserve surrounding anatomical context, and zero-pad the cropped volume to a fixed spatial size of $96 \times 256 \times 384$ (depth $\times$ height $\times$ width).

\textbf{3D Bounding Box Extraction.}
Given the processed segmentation masks, we extract 3D bounding boxes
using two strategies depending on the anatomical category. For normal organs, we compute the tight axis-aligned bounding box enclosing all voxels of each label. For lesion classes (\textit{e.g.}, kidney lesion, liver lesion, and pancreatic lesion), where multiple spatially disjoint instances may coexist in a single volume, we perform 3D connected component analysis with 26-connectivity to identify individual lesion instances. Connected components with fewer than 10 voxels are discarded as noise, and only the three largest instances (by voxel count) are retained for each lesion class. A per-instance bounding box is then computed for each retained component. All bounding box coordinates are normalized to $[0, 1]$ relative to the volume dimensions before being used as grounding supervision.

\textbf{2D Bounding Box Extraction.}
All 2D chest X-ray images are resized to $448 \times 448$ pixels. We extract disease-level bounding boxes from each dataset's native annotation format: for REFLACX, we parse the per-radiologist ellipse annotations and the chest region bounding box; for VinDr-CXR, we select annotations from the highest-priority radiologist with fallback to alternative annotators when the primary is absent, and use the consensus labels for the test set; for PadChest, we convert the normalized coordinates in the grounding annotations to pixel space. Across all datasets, bounding boxes belonging to the same disease category within a single image are aggregated into one sample, with duplicate boxes removed. All bounding box coordinates are then normalized to $[0, 1]$ relative to the resized image dimensions.

\subsection{Model Architecture Details}
\label{sec:appendix-model}
Our model adopts a dual-encoder multimodal large language model architecture to jointly process 2D radiographic images and 3D CT volumes. For 2D inputs (\textit{e.g.}, chest X-rays), we use the SigLIP-NaViT encoder from VideoLLaMA3~\citep{zhang2025videollama}, whose patch features are projected into the LLM embedding space via a two-layer MLP with GELU activation. For 3D CT volumes, we employ a convolutional encoder based on the PlainConvUNet architecture from nnU-Net~\citep{isensee2021nnu}. Features from the deepest encoder stage are projected via a $1\times1\times1$ convolution, flattened into a 1D token sequence, and mapped into the LLM space through a two-layer MLP projector. The 3D encoder is initialized from a pretrained segmentation model to leverage learned volumetric representations. We use Qwen3-VL~\citep{bai2025qwen3} as the language backbone, with 4B and 8B variants explored in our experiments. Visual tokens from either encoder are substituted into the corresponding placeholder positions in the input sequence, enabling the LLM to jointly attend to visual and textual information in a unified autoregressive framework.

\subsection{Public Benchmarks Details}
\label{sec:appendix-public_bench}
To verify the perceptual capabilities of our model, we jointly fine-tune the Stage~4 checkpoint on all public benchmarks and evaluate on their respective test sets. Table~\ref{tab:benchmark_datasets} summarizes the benchmarks used, covering both 2D and 3D medical VQA tasks
with open-ended, closed-ended, and multi-choice answer formats.
  
\begin{table}[t]
\centering
\caption{Statistics of downstream evaluation benchmarks.}
\label{tab:benchmark_datasets}
\resizebox{0.75\columnwidth}{!}{%
\begin{tabular}{@{}llccl@{}}
\toprule
\textbf{Modality} & \textbf{Dataset} & \textbf{Train} & \textbf{Test} & \textbf{Answer Type} \\
\midrule
\multirow{5}{*}{\textit{2D}}
& SLAKE~\citep{liu2021slake}  & 9,835   & 2,094   & Open + Closed \\
& OmniMedVQA~\citep{hu2024omnimedvqa} & 142,410 & 35,581  & Multi-Choice  \\
& PMC-VQA~\citep{zhang2023pmc}  & 152,603 & 33,430  & Multi-Choice  \\
& RadFig~\citep{yamagishi2025radfig}  & 228,155 & 10,139  & Multi-Choice  \\
\midrule
\multirow{3}{*}{\textit{3D}}
& 3D-Rad~\citep{3drad} & 134,701 & 33,712  & Open + Closed \\
& DeepTumor~\citep{chen2026deeptumorvqa} & 428,050 & 10,000  & Multi-Choice  \\
& CT-Spatial-VQA~\citep{monon2026ct-spatial} & -- & 8,953   & Open   \\
\bottomrule
\end{tabular}%
}
\end{table}
  
\subsection{Training Details}
\label{sec:appendix-training_details}
We adopt a four-stage curriculum training strategy, progressively introducing tasks of increasing complexity. Table~\ref{tab:training_details} summarizes the training configuration of each stage for both the 4B and 8B model variants.
All stages are trained on 32$\times$H20 GPUs with DeepSpeed ZeRO-Stage~1, gradient checkpointing, and BF16 mixed precision. To mitigate catastrophic forgetting of previously acquired capabilities, we employ a data replay strategy that mixes a subset of earlier-stage data into subsequent training stages. In Stage~3, we replay 20\% of Stage~2 data while training on the full anomaly detection dataset. In Stage~4, we further replay 10\% of Stage~2 and 20\% of Stage~3 data alongside the report generation data. For the anomaly detection subset in Stage~3 replay, we apply disease-level balanced sampling to maintain a 1:1 positive-to-negative ratio for each disease category. When the current-stage data is smaller than the replayed data, we upsample it via cyclic repetition to match the total volume.

\begin{table*}[t]
\centering
\caption{Training hyperparameters for each stage. ``--'' indicates the module is frozen. BS: batch size. LR: learning rate. Optim.: optimizer. Sched.: learning rate scheduler.}
\label{tab:training_details}
\resizebox{\textwidth}{!}{%
\begin{tabular}{@{}cccccccccc@{}}
\toprule
\textbf{Model} & \textbf{Stage} & \textbf{Data Size} & \textbf{Epochs} & \textbf{Global BS} & \textbf{Per-device BS} &
\textbf{LLM LR} & \textbf{Projector LR} & \textbf{Encoder LR} & \textbf{Optim. / Sched. / Warmup} \\
\midrule
\multirow{4}{*}{4B}
& 1 & 423K & 2 & 1024 & 8 & --   & 1e-3 & 1e-5 & AdamW / Cosine / 0.03 \\
& 2 & 4.56M & 1 & 1024 & 8 & 1e-5 & 1e-5 & 2e-6 & AdamW / Cosine / 0.03 \\
& 3 & 3.93M & 1 & 1024 & 8 & 1e-5 & 1e-5 & 2e-6 & AdamW / Cosine / 0.03 \\
& 4 & 2.12M & 2 & 1024 & 8 & 1e-5 & 1e-5 & 2e-6 & AdamW / Cosine / 0.03 \\
\midrule
\multirow{4}{*}{8B}
& 1 & 423K & 2 & 512 & 8 & --   & 1e-3 & 1e-5 & AdamW / Cosine / 0.03 \\
& 2 & 4.56M & 1 & 512 & 4 & 1e-5 & 1e-5 & 2e-6 & AdamW / Cosine / 0.03 \\
& 3 & 3.93M & 1 & 512 & 4 & 1e-5 & 1e-5 & 2e-6 & AdamW / Cosine / 0.03 \\
& 4 & 2.12M & 2 & 512 & 4 & 1e-5 & 1e-5 & 2e-6 & AdamW / Cosine / 0.03 \\
\bottomrule
\end{tabular}%
}
\end{table*}

\section{Evaluation Metrics}
\label{sec:appendix-metrics}

\subsection{Prompt-based LLM Evaluation for Attribute Judgment}
\label{sec:appendix-attribute_eval_prompt}
\begin{tcolorbox}[
    colback=black!5,  
    colframe=black!75, 
    title=Prompt Template for Attribute Judgment LLM Evaluation, 
    fonttitle=\bfseries,
    arc=2mm, 
    breakable,
]
\noindent
Please adhere strictly to the following rules: \\

\textbf{Score 1 (Correct / Fully Acceptable):} \\
The predict answer is fully correct and semantically equivalent to the reference answer, or provides a correct, more detailed version of it.

\begin{enumerate}
    \item \textbf{Exact or Semantic Match:} The answer is identical or uses different wording (synonyms, rephrasing) to convey the exact same meaning. \\
    Example: Ref: not pathological, Pred: not pathologically enlarged.

    \item \textbf{Correct but More Specific (Acceptable):} The predict answer is a correct, but more specific, instance of the reference answer. \\
    Example: Ref: viral pneumonias, Pred: Covid-19 viral pneumonia.

    \item \textbf{Contains Correct Additional Detail (Acceptable):} The predict answer includes all key information from the reference and adds other correct, relevant details. \\
    Example: Ref: sliding type, Pred: sliding type, Type I.

    \item \textbf{Numerical Answers:} When both answers are numbers, the predict answer matches exactly or is within a very close tolerance (e.g., $\pm$10\%) of the reference answer. \\
    Example: Ref: 10 mm, Pred: 9.5 mm.

    \item \textbf{Multi-dimensional Size Measurements:} When the reference answer contains a multi-dimensional measurement (e.g., ``A $\times$ B cm'') and the predict answer provides only a single dimension, use the largest dimension of the reference as the basis for comparison, then apply the standard numerical tolerance rules. \\
    Example: Ref: 1.9 $\times$ 0.5 cm, Pred: 1.8 cm $\rightarrow$ compare 1.8 cm against 1.9 cm (the largest dimension) $\rightarrow$ difference is $\sim$5\%, within $\pm$10\% $\rightarrow$ Score 1. \\
    Example: Ref: 3.0 $\times$ 1.2 cm, Pred: 2.8 cm $\rightarrow$ compare 2.8 cm against 3.0 cm $\rightarrow$ difference is $\sim$7\%, within $\pm$10\% $\rightarrow$ Score 1. \\
    Example: Ref: 3.0 $\times$ 1.2 cm, Pred: 1.1 cm $\rightarrow$ compare 1.1 cm against 3.0 cm (largest) and 1.2 cm (second) $\rightarrow$ 1.1 cm is close to 1.2 cm (second dimension), not the largest $\rightarrow$ Score 0.
\end{enumerate}

\textbf{Score 0.5 (Partially Correct):} \\
The predict answer is on the right track but is flawed by being incomplete or too general.

\begin{enumerate}
    \item \textbf{Correct but Incomplete:} The predict answer provides correct information but omits some key elements from the reference answer. \\
    Example: Ref: ground-glass, consolidation, Pred: ground-glass. \\
    Example: Ref: right lung middle lobe, left lingular segment, Pred: right lung middle lobe.

    \item \textbf{Correct but Overly General:} The predict answer is a correct but less specific version of the reference answer. It captures the essence but loses important detail. \\
    Example 1 (Qualitative): Ref: 6 mm, Pred: millimetric. \\
    Example 2 (Location): Ref: left lung lower lobe superior segment, right lung lower lobe, Pred: both lungs, more prominently in the lower lobes.

    \item \textbf{Numerical Answers:} The predict answer is numerically in the same ballpark (same order of magnitude) but outside the strict tolerance for a score of 1. It captures part of the quantitative information but is imprecise. \\
    Example: Ref: 8 mm, Pred: 10 mm. (Correct order of magnitude, 30\% $>$ difference $>$ 10\%).

    \item \textbf{Multi-dimensional Size Measurements:} When the reference contains a multi-dimensional measurement and the predict answer provides only a single dimension that matches the largest dimension within the 0.5-score tolerance range (10\%$\sim$30\% difference), assign Score 0.5. \\
    Example: Ref: 2.0 $\times$ 0.8 cm, Pred: 1.6 cm $\rightarrow$ compare against 2.0 cm $\rightarrow$ difference is 20\%, within 10\%$\sim$30\% range $\rightarrow$ Score 0.5.

    \item \textbf{Correct Laterality but Wrong Lobe:} The predict answer correctly identifies the side (left/right) but specifies a wrong lobe or segment within that side. Since the laterality is correct but the intra-lung localization is wrong, this is partially correct. \\
    Example 1: Ref: right lung upper lobe, Pred: right lung lower lobe $\rightarrow$ correct side (right), wrong lobe $\rightarrow$ Score 0.5. \\
    Example 2: Ref: left lung lower lobe, Pred: left lung upper lobe $\rightarrow$ correct side (left), wrong lobe $\rightarrow$ Score 0.5. \\
    Example 3: Ref: right lung middle lobe, Pred: right lung lower lobe $\rightarrow$ correct side (right), wrong lobe $\rightarrow$ Score 0.5. \\
    \textit{Note: This rule only applies when the laterality is unambiguously correct and only the lobe/segment is wrong. If the side itself is wrong or expanded to bilateral, apply Score 0 rules instead.}
\end{enumerate}

\textbf{Score 0 (Incorrect):} \\
The predict answer is factually wrong, irrelevant, or fails to answer the question.

\begin{enumerate}
    \item \textbf{Contradiction:} It directly contradicts the reference answer. \\
    Example: Ref: pathological, Pred: not pathologically enlarged.

    \item \textbf{Wrong Information:} It provides a completely different piece of information. \\
    Example: Ref: atelectasis, Pred: infectious process.

    \item \textbf{Hallucination / Irrelevant:} It provides information not supported by the context or states that an answer cannot be determined when the reference provides one. \\
    Example: Ref: focal, Pred: The report does not provide this information.

    \item \textbf{Numerical Answers:} The predict answer is of a different order of magnitude or significantly incorrect (difference $>$ 30\%). \\
    Example: Ref: 10 mm, Pred: 1 mm or 50 mm.

    \item \textbf{Wrong Laterality or Wrong Side:} The reference specifies a particular side (e.g., left), but the predict answer states the opposite side (e.g., right) or expands it to bilateral/both sides. Changing or broadening the laterality is always incorrect and must be scored 0. \\
    Example 1: Ref: left lung, Pred: right lung. (opposite side $\rightarrow$ Score 0) \\
    Example 2: Ref: left lung, Pred: bilateral / both lungs. (unilateral expanded to bilateral $\rightarrow$ Score 0) \\
    Example 3: Ref: right kidney, Pred: both kidneys. (unilateral expanded to bilateral $\rightarrow$ Score 0)

    \item \textbf{Multi-dimensional Size Measurements:} When the predict answer provides only a single dimension that does not match the largest dimension of the reference (difference $>$ 30\%), assign Score 0. \\
    Example: Ref: 3.0 $\times$ 1.2 cm, Pred: 1.1 cm $\rightarrow$ does not match largest dimension (3.0 cm), difference $>$ 30\% $\rightarrow$ Score 0.
\end{enumerate}

\textbf{Instruction} \\
You are an expert evaluator for medical question answering systems. Your task is to score a predict answer against a reference answer for a given question. You must evaluate the semantic accuracy and completeness of the predict answer with the above scoring criteria. Your final output must consist of a score from the set \{0, 0.5, 1\}. You need to think first, and then give the final score. The final score should be enclosed in \texttt{<score>} and \texttt{</score>}. \\
Example: your thinking process. \texttt{<score>} 1 or 0.5 or 0 \texttt{</score>} \\

\textbf{Input} \\
Question: \texttt{<question>} \\
Reference Answer: \texttt{<ref>} \\
Predict Answer: \texttt{<prd>}

\end{tcolorbox}

\subsection{Grounding Metrics}
\label{sec:appendix-grounding_metrics}
We evaluate the spatial grounding capability of our model using both 2D and 3D bounding box predictions, depending on the imaging modality. For 2D modalities (\emph{e.g.}, X-ray), each bounding box is represented as $\mathbf{b} = (x_1, y_1, x_2, y_2)$, where $(x_1, y_1)$ and $(x_2, y_2)$ denote the top-left and bottom-right corners, respectively. For 3D volumetric modalities (\emph{e.g.}, CT), bounding boxes are extended to $\mathbf{b} = (x_1, y_1, z_1, x_2, y_2, z_2)$, where the additional coordinates specify the extent along the axial (depth) dimension.

\textbf{Intersection over Union (IoU).}
Given a predicted bounding box $\mathbf{b}^p$ and a ground-truth bounding box $\mathbf{b}^g$, the 2D IoU is computed as:
\begin{equation}
    \mathrm{IoU}_{\text{2D}} = \frac{|\mathbf{b}^p \cap \mathbf{b}^g|}{|\mathbf{b}^p \cup \mathbf{b}^g|} = \frac{A_{\text{inter}}}{A^p + A^g - A_{	ext{inter}}},
\end{equation}
where $A^p$ and $A^g$ denote the areas of the predicted and ground-truth boxes, and the intersection area is given by:
\begin{equation}
    A_{\text{inter}} = \max(0,\; x_2^{\min} - x_1^{\max})\times \max(0,\; y_2^{\min} - y_1^{\max}),
\end{equation}
with $x_1^{\max} = \max(x_1^p, x_1^g)$, $x_2^{\min} = \min(x_2^p, x_2^g)$, and analogously for $y$.

For 3D bounding boxes, the IoU is naturally extended to volumetric overlap:
\begin{equation}
    \mathrm{IoU}_{\text{3D}} = \frac{V_{	ext{inter}}}{V^p + V^g - V_{	ext{inter}}},
\end{equation}
where the intersection volume is:
\begin{equation}
    V_{\text{inter}} = \prod_{d \in \{x, y, z\}} \max\!\Big(0,\; \min(d_2^p, d_2^g) - \max(d_1^p, d_1^g)\Big),
\end{equation}
and $V^p = \prod_{d \in \{x,y,z\}} (d_2^p - d_1^p)$, $V^g = \prod_{d \in \{x,y,z\}} (d_2^g - d_1^g)$ are the volumes of the predicted and ground-truth boxes, respectively.

\textbf{Greedy Matching.}
When multiple bounding boxes are present in a single sample, we employ a greedy matching strategy to establish correspondences between predictions and ground-truth boxes. Specifically, we first compute the full IoU matrix $\mathbf{M} \in \mathbb{R}^{N_g \times N_p}$, where $N_g$ and $N_p$ denote the number of ground-truth and predicted boxes, respectively. All $(i, j)$ pairs are sorted by $\mathbf{M}_{ij}$ in descending order, and matches are greedily assigned such that each ground-truth and predicted box is matched at most once. A match is accepted only if $\mathbf{M}_{ij} \geq \tau$, where $\tau$ is a predefined IoU threshold (set to $\tau = 0.3$ in our experiments).

\textbf{Detection Metrics.}
Based on the matching results, we compute Precision, Recall, and F1-score as:
\begin{equation}
    \mathrm{Precision} = \frac{\mathrm{TP}}{\mathrm{TP} + \mathrm{FP}}, \quad
    \mathrm{Recall} = \frac{\mathrm{TP}}{\mathrm{TP} + \mathrm{FN}}, \quad
    \mathrm{F1} = \frac{2 \cdot \mathrm{Precision} \cdot \mathrm{Recall}}{\mathrm{Precision} + \mathrm{Recall}},
\end{equation}
where $\mathrm{TP}$, $\mathrm{FP}$, and $\mathrm{FN}$ denote the number of true positives (successfully matched pairs), false positives (unmatched predictions), and false negatives (unmatched ground-truth boxes), respectively.

\textbf{Mean IoU (mIoU).}
To quantify localization accuracy, we report the mean IoU (mIoU) across all evaluated samples. For single-box samples, the IoU between the predicted and ground-truth box is always included regardless of the threshold $\tau$, providing a continuous measure of localization quality. For multi-box samples, only the IoU values from successfully matched pairs (with $\mathrm{IoU} \geq \tau$) are included. The overall mIoU is then computed as:
\begin{equation}
    \mathrm{mIoU} = \frac{1}{|\mathcal{S}|} \sum_{s \in \mathcal{S}} \mathrm{IoU}_s,
\end{equation}
where $\mathcal{S}$ is the set of all valid IoU values collected according to the above strategy.

\subsection{Clinical Evaluation Metrics}
\label{sec:appendix-clinical_metrics}
\textbf{Why is AUC used as the evaluation metric for anomaly detection?}
Many previous anomaly-detection studies primarily report accuracy (ACC) as
the evaluation metric~\citep{3drad}. However, ACC depends strongly on both the
class distribution and the decision threshold, and may therefore overestimate
diagnostic performance under severe class imbalance.

\begin{figure*}[t]
    \centering
    \includegraphics[width=1\linewidth]{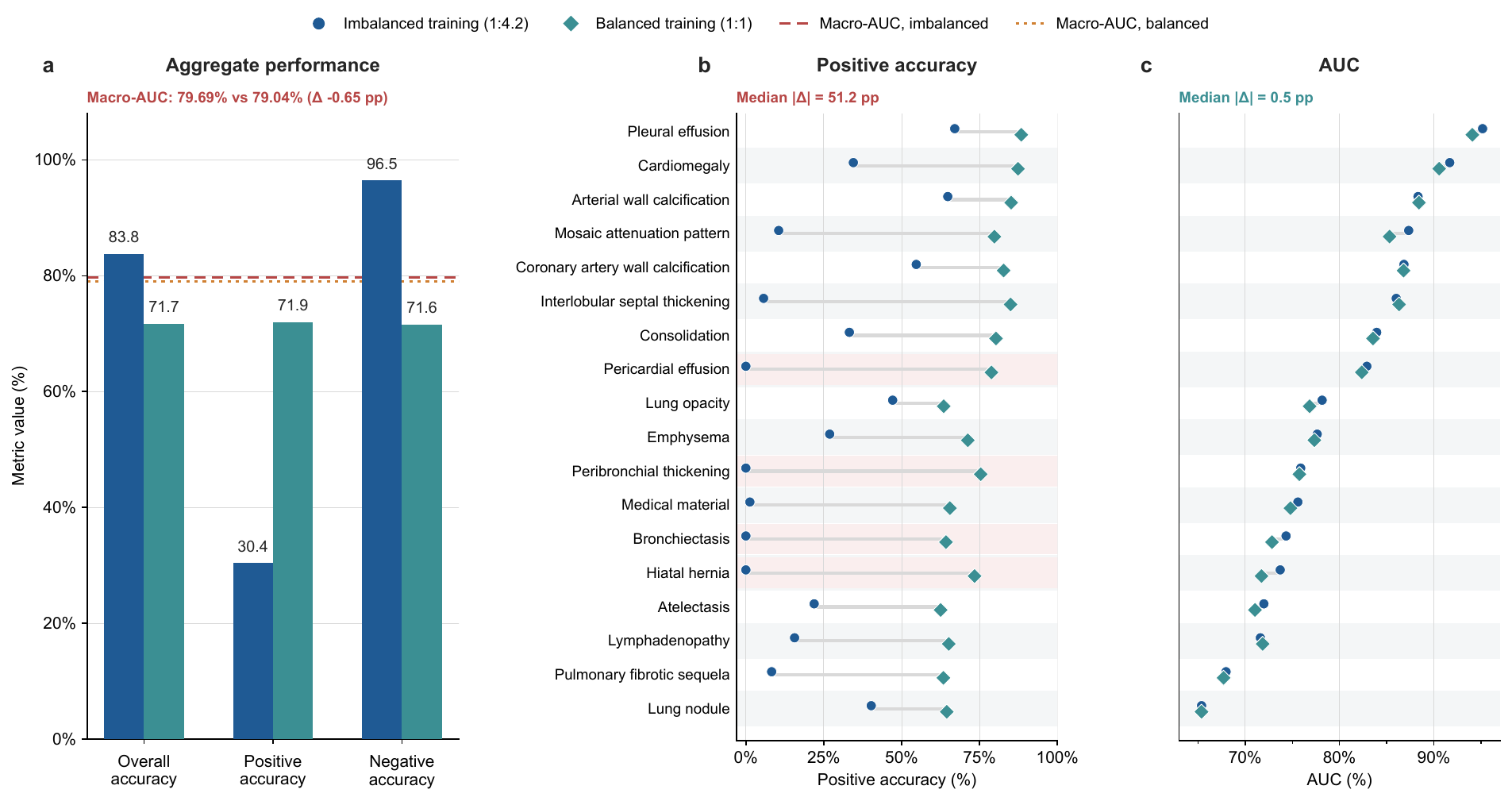}
    \caption{
        \textbf{AUC provides a more reliable evaluation of anomaly-detection
        ability under class imbalance.}
        \textbf{a}, Aggregate performance under imbalanced ($1{:}4.2$) and
        balanced ($1{:}1$) training distributions. Although balancing
        substantially improves positive accuracy, the overall ACC decreases
        because fewer samples belong to the majority negative class. The two
        horizontal lines indicate the corresponding macro-average AUC values.
        \textbf{b}, Disease-level positive accuracy changes markedly after
        class balancing, with a median absolute difference of $51.2$
        percentage points (pp). Light-red rows denote diseases for which the
        model produces no positive predictions under imbalanced training.
        \textbf{c}, Disease-level AUC remains comparatively stable across the
        two training distributions, with a median absolute difference of only
        $0.5$~pp. Circles and diamonds represent models trained with the
        imbalanced and balanced distributions, respectively.
    }
    \label{fig:auc_vs_acc}
\end{figure*}

Taking the CT-RATE dataset as an example, we formulate 18 disease-specific
questions for each patient. Constructing the training set according to the
natural label distribution results in a positive-to-negative ratio of
approximately $1{:}4.2$. Under this distribution, a model can achieve high ACC
by favoring the majority negative class, even when it rarely identifies
positive cases. As shown in Figure~\ref{fig:auc_vs_acc}~a, the model trained with the original imbalanced distribution achieves an overall ACC of $83.75\%$. However, its
positive accuracy is only $30.38\%$, compared with a negative accuracy of
$96.45\%$. For four diseases, the model produces no positive predictions,
resulting in an F1 score of zero. The apparently high ACC therefore primarily
reflects the dominance of negative samples rather than reliable diagnostic
ability.

We further reconstruct the training set using a balanced $1{:}1$
positive-to-negative ratio. After balancing, positive accuracy increases from
$30.38\%$ to $71.92\%$, and the disease-level F1 scores improve, including
recovery from zero for the four previously undetected diseases. Nevertheless,
overall ACC decreases from $83.75\%$ to $71.65\%$. This counter-intuitive result
shows that ACC may penalize a model for becoming more sensitive to clinically
relevant positive cases. By contrast, the macro-average area under the receiver operating characteristic curve (AUC) remains nearly unchanged: $79.69\%$ under
imbalanced training and $79.04\%$ after balancing, corresponding to a
difference of only $0.65$~pp. The disease-level comparison further shows that
the median absolute change in positive accuracy is $51.2$~pp
(Figure~\ref{fig:auc_vs_acc}~b), whereas the corresponding change in AUC is only
$0.5$~pp (Figure~\ref{fig:auc_vs_acc}~c).

Unlike ACC, which evaluates binary predictions at a single operating
threshold, AUC measures how consistently positive cases are ranked above
negative cases across possible thresholds. It is therefore less sensitive to
a particular decision threshold and to the class prevalence of the evaluation
set. These results do not imply that AUC is mathematically invariant to the
training distribution; rebalancing may still alter the learned ranking
function. Rather, they demonstrate that, in our experiments, the underlying
discriminative ability remains stable while the threshold-dependent binary
decisions change substantially. We use AUC as the primary metric for
anomaly detection.

\textbf{Area Under the ROC Curve (AUC).}
For binary clinical diagnosis tasks (\emph{e.g.}, determining whether a specific disease is present), we frame the evaluation as a binary classification problem. The model produces a confidence score $p \in [0, 1]$ representing the estimated probability that the target pathology is present (``yes''). Given a decision threshold $t$, samples with $p \geq t$ are classified as positive. By varying $t$ across all possible values, we obtain the Receiver Operating Characteristic (ROC) curve, which plots the True Positive Rate (TPR, \emph{i.e.}, Sensitivity) against the False Positive Rate (FPR, \emph{i.e.}, $1 - \text{Specificity}$):
\begin{equation}
    \mathrm{TPR}(t) = \frac{\mathrm{TP}(t)}{\mathrm{TP}(t) + \mathrm{FN}(t)}, \quad
    \mathrm{FPR}(t) = \frac{\mathrm{FP}(t)}{\mathrm{FP}(t) + \mathrm{TN}(t)},
\end{equation}
where $\mathrm{TP}$, $\mathrm{FP}$, $\mathrm{TN}$, and $\mathrm{FN}$ denote the number of true positives, false positives, true negatives, and false negatives at threshold $t$, respectively. The AUC is then defined as the area under this curve:
\begin{equation}
    \mathrm{AUC} = \int_0^1 \mathrm{TPR}\big(\mathrm{FPR}^{-1}(x)\big) \, dx.
\end{equation}
An AUC of $1.0$ indicates perfect discrimination, while $0.5$ corresponds to random chance. Importantly, AUC is a threshold-independent metric, making it robust to class imbalance --- a common characteristic of clinical datasets where positive (diseased) cases are often significantly outnumbered by negative (healthy) ones.

To derive clinically interpretable operating points, we additionally adopt the Youden's index~\citep{youden1950index} to select the optimal binarization threshold:
\begin{equation}
    t^{*} = \arg\max_{t} \Big\{ \mathrm{TPR}(t) - \mathrm{FPR}(t) \Big\} = \arg\max_{t} \Big\{ \mathrm{Sensitivity}(t) + \mathrm{Specificity}(t) - 1 \Big\}.
\end{equation}
At the optimal threshold $t^{*}$, we report Sensitivity and Specificity as secondary metrics:
\begin{equation}
    \mathrm{Sensitivity} = \frac{\mathrm{TP}}{\mathrm{TP} + \mathrm{FN}}, \quad
    \mathrm{Specificity} = \frac{\mathrm{TN}}{\mathrm{TN} + \mathrm{FP}}.
\end{equation}
All metrics are computed per disease category within each imaging modality, and we report the macro-averaged AUC across diseases as the primary aggregate metric.

\textbf{Multiple-Choice Accuracy (ACC).}
For clinical knowledge questions presented in multiple-choice format, we evaluate accuracy by comparing the model's selected option against the ground-truth answer. For single-answer questions, a prediction is scored $1$ if it matches the correct option and $0$ otherwise. For multi-answer questions, we adopt a partial-credit scoring scheme: if the predicted option set $\mathcal{P}$ contains any incorrect option (\emph{i.e.}, $\mathcal{P} \not\subseteq \mathcal{G}$, where $\mathcal{G}$ is the ground-truth set), the score is $0$; otherwise, the score is proportional to the number of correctly selected options:
\begin{equation}
    \mathrm{Score} =
    \begin{cases}
        0, & \text{if } \mathcal{P} \not\subseteq \mathcal{G}, \\[4pt]
        \displaystyle \frac{|\mathcal{P} \cap \mathcal{G}|}{|\mathcal{G}|}, & \text{otherwise}.
    \end{cases}
\end{equation}

The overall accuracy is the mean score across all samples.

\subsection{Report Generation Evaluation Metrics.}
\label{sec:appendix-mrg_metrics}
Unlike most prior works that adopt NLG metrics (\emph{e.g.}, BLEU~\citep{papineni2002bleu}, ROUGE~\citep{lin2004rouge}) for report evaluation, we note that such $n$-gram overlap metrics are ill-suited for clinical settings: they treat all tokens equally, thus failing to capture the impact of negation words (\emph{e.g.}, ``no evidence of'') that reverse diagnostic conclusions, and penalize valid medical synonyms (\emph{e.g.}, ``heart enlargement'' \emph{vs.} ``cardiomegaly'') despite identical clinical semantics. Instead, we adopt \textbf{Clinical Efficacy (CE) F1} and \textbf{RaTEScore}~\citep{zhao2024ratescore}, which evaluate factual correctness at the pathology-label and entity levels, respectively.

\textbf{Clinical Efficacy F1.}
We adopt Clinical Efficacy (CE) metrics to assess whether the generated report conveys the same clinical findings as the reference, by comparing pathology label vectors rather than surface text. The ground-truth labels are obtained directly from dataset annotations. For CT-RATE~\citep{ct-rate}, 18 thoracic pathology categories are provided; for MIMIC-CXR~\citep{mimic}, 14 pathology labels are extracted by CheXbert~\citep{smit2020chexbert}. Predicted labels are extracted from the generated reports using a RadBERT-based classifier~\citep{yan2022radbert} (for CT-RATE) or CheXbert (for MIMIC-CXR). We report Macro F1, Micro F1, and sample-wise F1 to measure the agreement between ground-truth and predicted label vectors.

\textbf{RaTEScore.}
RaTEScore~\citep{zhao2024ratescore} is an entity-aware metric that extracts medical entities from both the reference and candidate reports via a fine-tuned NER model, encodes them with BioLORD~\citep{remy2024biolord}, and computes entity-level precision and recall through type-weighted cosine similarity matching with negation penalties. The final score is the harmonic mean of precision and recall.

\section{Additional Experimental Results}
\label{sec:appendix-results}

\subsection{Ablation Studies}
\label{sec:appendix-ablation}

\textbf{Impact of Perceptual Curriculum Learning on 2D X-ray Report Generation.}
\begin{figure}[t]
\centering
\includegraphics[width=0.75\linewidth]{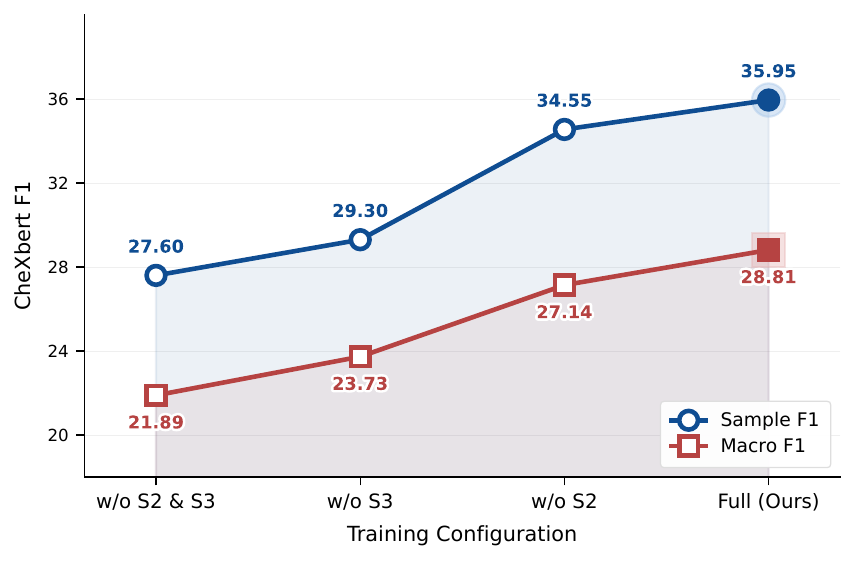}
\caption{Ablation study on the impact of intermediate training stages on 2D chest X-ray report generation performance (MIMIC-CXR). CheXbert F1 metrics are reported.}
\label{fig:ablation_2d}
\end{figure}
Since publicly available 2D medical benchmarks typically aggregate multiple imaging modalities (\eg, X-ray, MRI, CT), isolating the contribution of each training stage to a specific modality is challenging. Given that our 2D training data is predominantly composed of chest X-rays, we conduct the ablation study on MIMIC-CXR to ensure a consistent and controlled evaluation, as shown in Figure~\ref{fig:ablation_2d}. We selectively remove Stage~2 and Stage~3 from the training pipeline. Removing both stages yields the lowest performance (Sample F1: 27.60), confirming that directly fine-tuning from alignment is insufficient. Adding Stage~2 alone improves Sample F1 to 29.30 (+1.70), while adding Stage~3 alone yields a larger gain (34.55, +6.95), indicating that diagnostic reasoning contributes more than perceptual grounding. The full model achieves the best result (Sample F1: 35.95, Macro F1: 28.81), validating that progressively building from perception to diagnosis enables increasingly effective report generation, and confirming that perceptual curriculum learning is also effective for 2D X-ray.

\begin{figure}[t]
\centering
\includegraphics[width=0.75\linewidth]{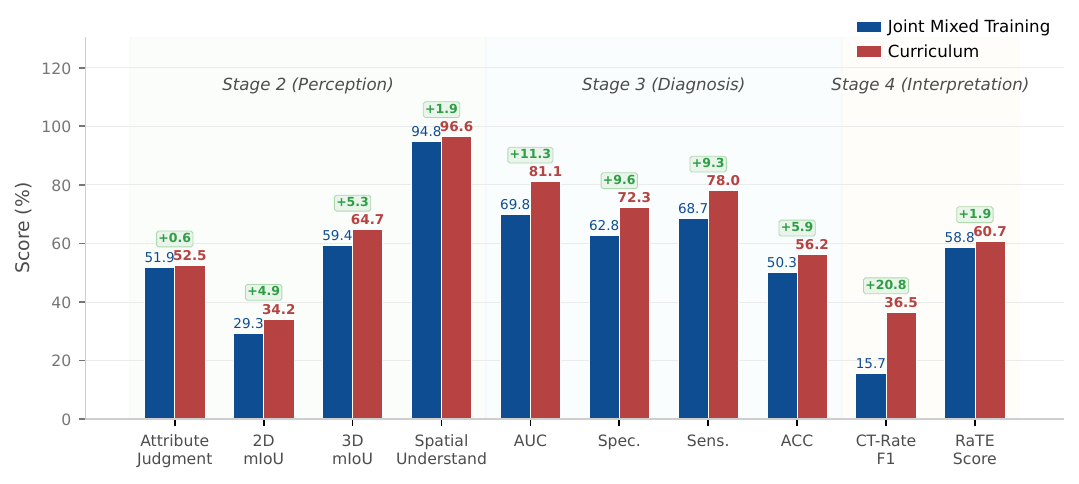}
\caption{Ablation study on the impact of curriculum learning. Performance comparison across ten metrics spanning perception, diagnosis, and interpretation. Curriculum learning consistently outperforms joint mixed training.}
\label{fig:curriculum_vs_joint}
\end{figure}

\textbf{Curriculum Learning vs.\ Joint Mixed Training.}
To validate the necessity of our progressive training strategy, we compare the four-stage curriculum learning pipeline against a joint mixed training baseline that trains on all tasks simultaneously without stage separation. As shown in Figure~
\ref{fig:curriculum_vs_joint}, curriculum learning consistently outperforms joint mixed training across all ten evaluation metrics spanning perception, diagnosis, and interpretation. The most pronounced gains appear in CT-Rate F1 (+20.8) and AUC (+11.3), indicating that staged training is especially beneficial for tasks requiring complex clinical reasoning. Even for perception-level metrics such as 3D mIoU (+5.3) and 2D mIoU (+4.9), the progressive strategy yields notable improvements, suggesting that learning lower-level visual grounding before higher-level reasoning helps the model develop more robust spatial representations. These results confirm that simply mixing all training data cannot substitute for a carefully designed curriculum, and that progressively building from alignment through perception and diagnosis to interpretation is essential for achieving strong performance across the full spectrum of medical vision-language tasks.

\subsection{Full Experimental Results}
\label{sec:appendix-result_details}

\subsubsection{Spatial Grounding}
\textbf{Spatial grounding performance on lesion detection.}
Most existing baseline models lack the ability to localize lesions and organs within 3D volumes, as their bounding box outputs are typically 2D coordinate quadruples, which are incompatible with our 3D bounding boxes and thus preclude direct metric comparison. In Table~\ref{tab:lesion_bbox}, we report the lesion localization performance of RadSight. Compared to organ localization, lesion samples are considerably scarce, and a single patient may harbor multiple small lesions scattered across the volume, making it inherently challenging for the model to pinpoint the exact location of each lesion with limited training data. As a result, the overall mIoU appears relatively low. Nevertheless, for lesions with relatively larger volumes, RadSight is able to accurately localize them, as illustrated in the qualitative visualizations in Figure~\ref{fig:case_3d_lesion}.
\begin{table}[h]
\centering
\caption{Spatial grounding performance on lesion detection.}
\label{tab:lesion_bbox}
\begin{tabular}{lc}
\toprule
Disease & mIoU \\
\midrule
Pancreatic lesion & 8.30 \\
Kidney lesion     & 7.60 \\
Liver lesion      & 2.60 \\
\bottomrule
\end{tabular}
\end{table}

\begin{table*}[t]
\centering
\caption{Per-disease performance of abnormality detection on the combined X-ray and CT test set. We report the area under the ROC curve (AUC), sensitivity (Sen), and specificity (Spe) for each of the 82 evaluated diseases. The global macro-averaged AUC is 81.10\%.}
\label{tab:auc_per_disease}
\vspace{2pt}
\resizebox{\textwidth}{!}{%
\begin{tabular}{lccc|lccc}
\toprule
\textbf{Disease} & \textbf{AUC} & \textbf{Sen} & \textbf{Spe} & \textbf{Disease} & \textbf{AUC} & \textbf{Sen} & \textbf{Spe} \\
\midrule
Abscess                        & 86.44 & 83.33 & 76.36 & Kidney Lesion                  & 92.09 & 85.71 & 82.65 \\
Anasarca                       & 92.99 & 89.84 & 82.63 & Kidney Stone                   & 71.39 & 52.00 & 90.64 \\
Aortic Aneurysm                & 82.93 & 100.0 & 60.75 & Kyphosis                       & 76.57 & 85.71 & 56.71 \\
Aortic Valve Calcification     & 90.01 & 87.63 & 79.11 & Liver Lesion                   & 85.23 & 72.15 & 84.91 \\
Appendicitis                   & 89.70 & 83.94 & 83.14 & Lung Lesion                    & 68.23 & 85.48 & 42.51 \\
Arterial Wall Calcification    & 94.39 & 91.76 & 85.65 & Lung Nodule                    & 69.56 & 77.48 & 51.13 \\
Ascites                        & 84.15 & 82.09 & 70.63 & Lung Opacity                   & 72.85 & 66.45 & 66.44 \\
Atelectasis                    & 76.61 & 74.64 & 64.42 & Lymphadenopathy                & 74.35 & 70.02 & 66.78 \\
Atherosclerosis                & 87.09 & 81.06 & 78.12 & Medical Material               & 90.99 & 80.38 & 85.70 \\
Biliary Ductal Dilation        & 92.81 & 80.08 & 90.72 & Mosaic Attenuation Pattern      & 84.32 & 76.98 & 77.89 \\
Bladder Wall Thickening        & 73.19 & 67.20 & 69.49 & Osteopenia                     & 70.40 & 67.83 & 61.65 \\
Bowel Obstruction              & 93.87 & 86.26 & 90.19 & Osteophyte                     & 73.57 & 78.95 & 64.02 \\
Bronchiectasis                 & 73.31 & 77.46 & 57.01 & Ovarian Cyst                   & 82.83 & 82.54 & 66.91 \\
Cardiomegaly                   & 88.19 & 88.64 & 73.89 & Pancreas Lesion                & 82.53 & 82.73 & 68.38 \\
Cirrhosis                      & 97.34 & 95.88 & 92.58 & Pancreatic Atrophy             & 80.90 & 77.54 & 71.40 \\
Colitis                        & 80.53 & 85.25 & 67.24 & Pancreatic Cyst                & 71.87 & 52.05 & 80.19 \\
Colon Lesion                   & 98.85 & 95.00 & 97.68 & Pancreatomegaly                & 79.91 & 69.05 & 74.51 \\
Consolidation                  & 75.81 & 70.83 & 67.30 & Peribronchial Thickening       & 79.02 & 84.00 & 60.76 \\
Coronary Artery Disease        & 83.56 & 91.49 & 72.51 & Pericardial Effusion           & 83.63 & 66.42 & 85.89 \\
Coronary Calcification         & 89.89 & 83.73 & 80.57 & Pleural Effusion               & 94.14 & 90.10 & 84.26 \\
Degenerative Changes           & 65.41 & 62.50 & 60.89 & Pneumonia                      & 74.53 & 83.30 & 54.37 \\
Degenerative Disc Disease      & 67.88 & 95.65 & 30.72 & Pneumothorax                   & 81.25 & 83.54 & 65.03 \\
Diverticulitis                 & 88.59 & 82.17 & 78.56 & Post-operative                 & 89.18 & 78.58 & 85.39 \\
Diverticulosis                 & 78.69 & 75.31 & 68.68 & Prostatomegaly                 & 89.96 & 89.53 & 77.88 \\
Edema                          & 82.62 & 83.89 & 65.89 & Pulmonary Edema                & 87.06 & 83.13 & 75.13 \\
Emphysema                      & 81.46 & 86.91 & 62.73 & Pulmonary Embolism             & 64.95 & 60.47 & 59.61 \\
Enlarged Cardiomediastinum     & 67.92 & 79.47 & 49.29 & Pulmonary Fibrotic Sequela     & 68.67 & 68.62 & 60.69 \\
Fibrosis                       & 65.17 & 44.44 & 80.09 & Pulmonary Hypertension         & 80.87 & 69.23 & 77.65 \\
Fracture                       & 68.23 & 51.75 & 75.26 & Renal Atrophy                  & 88.77 & 85.71 & 78.15 \\
Free Air                       & 72.56 & 58.84 & 74.46 & Renal Calculus                 & 75.56 & 55.96 & 84.49 \\
Gallbladder Wall Thickening    & 89.23 & 95.24 & 71.02 & Renal Cyst                     & 82.81 & 73.93 & 75.27 \\
Gallstones                     & 90.00 & 74.44 & 90.85 & Renal Hypodensities            & 64.84 & 69.36 & 52.11 \\
Granuloma                      & 65.32 & 54.39 & 68.69 & Scoliosis                      & 68.93 & 87.38 & 40.69 \\
Hemangioma                     & 73.46 & 61.31 & 72.51 & Splenomegaly                   & 94.88 & 90.87 & 86.09 \\
Hematoma                       & 69.67 & 92.00 & 42.11 & Submucosal Edema               & 74.24 & 69.84 & 69.21 \\
Hepatic Steatosis              & 84.59 & 70.16 & 80.62 & Support Devices                & 88.66 & 78.47 & 85.65 \\
Hepatomegaly                   & 87.42 & 74.12 & 84.13 & Surgically Absent Gallbladder  & 97.15 & 97.06 & 93.06 \\
Hernia                         & 78.47 & 76.01 & 67.29 & Thrombosis                     & 77.18 & 68.00 & 71.92 \\
Hydronephrosis                 & 88.09 & 77.72 & 85.07 & Thyroid Nodule                 & 63.80 & 58.65 & 63.09 \\
Interlobular Septal Thickening & 85.73 & 81.45 & 75.56 & Uterine Fibroids               & 84.72 & 85.94 & 70.03 \\
Kidney Enlargement             & 90.81 & 87.81 & 76.73 & Varices                        & 94.97 & 94.87 & 82.90 \\
\midrule
\multicolumn{8}{c}{\textbf{Global Mean:\quad AUC = 81.10\quad Sen = 78.02\quad Spe = 72.35}} \\
\bottomrule
\end{tabular}%
}
\end{table*}

\textbf{Zero-Shot Spatial Grounding on Additional Benchmarks.}
Table~\ref{tab:btcv_amos_zeroshot} reports the zero-shot spatial grounding performance on BTCV~\citep{landman2015btcv} and AMOS-22~\citep{ji2022amos}, two widely-used abdominal organ segmentation benchmarks. Without any task-specific fine-tuning, our RadSight significantly outperforms M3D on both datasets, achieving 47.54 mIoU and 80.67 F1 on BTCV, and 44.57 mIoU and 75.62 F1 on AMOS-22, improving F1 by 51.47 points on BTCV and 25.89 points on AMOS-22. These results demonstrate the strong generalization
ability of our model for 3D anatomical localization in unseen datasets.
\begin{table}[h]
\centering
\caption{Zero-shot spatial grounding performance on BTCV and AMOS-22. F1 is computed at IoU threshold $\tau =
0.3$.}
\label{tab:btcv_amos_zeroshot}
\begin{tabular}{lcccc}
\toprule
Model & BTCV-mIoU & BTCV-F1 & AMOS-22-mIoU & AMOS-22-F1 \\
\midrule
M3D-Phi-3-4B  & 19.78 & 29.20 & 29.29 & 49.73 \\
RadSight-4B    & \textbf{47.54} & \textbf{80.67} & \textbf{44.57} & \textbf{75.62} \\
\bottomrule
\end{tabular}
\end{table}

\subsubsection{Report Generation}
\textbf{Report Generation on CT-Rate.}
Table~\ref{tab:ct_rate_report_gen} presents the report generation performance on CT-RATE, evaluated by Clinical Efficacy (CE) metrics using a RadBERT-based classifier. Our method achieves the highest F1 score of 36.5, outperforming the best baseline BTB3D by a large margin of 10.7. Notably, while prior methods such as CT2Rep and CT-CHAT attain competitive Precision, their Recall remains below 16, indicating a tendency to generate conservative reports that miss many positive findings. In contrast, our method achieves a substantially higher Recall of 32.0 while maintaining the best Precision of 45.3, demonstrating a better balance between diagnostic coverage and correctness. We provide the per-disease breakdown of our method in Table~\ref{tab:ct_rate_per_disease}.
\begin{table}[h]
\centering
\caption{Report generation performance on CT-RATE. Results of compared methods are from~\citep{hamamci2026btb3d}.}
\label{tab:ct_rate_report_gen}
\begin{tabular}{lccc}
\toprule
Method & Prec & Recall & F1 \\
\midrule
CT2Rep~\citep{hamamci2024ct2rep}  & 43.5 & 12.8 & 16.0 \\
CT-CHAT~\citep{hamamci2026ct-chat} & 45.0 & 15.8 & 18.4 \\
Merlin~\citep{merlin}  & 29.5 & 11.2 & 16.0 \\
BTB3D~\citep{hamamci2026btb3d}   & 26.0 & 26.0 & 25.8 \\
\midrule
Ours    & \textbf{45.3} & \textbf{32.0} & \textbf{36.5} \\
\bottomrule
\end{tabular}
\end{table}

\begin{table}[t]
\centering
\caption{Per-disease Clinical Efficacy results on CT-RATE.}
\label{tab:ct_rate_per_disease}
\begin{tabular}{lccc}
\toprule
\textbf{Disease} & \textbf{F1} & \textbf{Precision} & \textbf{Recall} \\
\midrule
Medical material               & 29.41 & 58.82 & 19.61 \\
Arterial wall calcification    & 70.00 & 76.69 & 64.39 \\
Cardiomegaly                   & 53.33 & 64.86 & 45.28 \\
Pericardial effusion           & 16.67 & 26.53 & 12.15 \\
Coronary artery wall calc.     & 65.65 & 69.30 & 62.37 \\
Hiatal hernia                  & 17.72 & 27.72 & 13.02 \\
Lymphadenopathy                & 23.16 & 42.00 & 15.99 \\
Emphysema                      & 40.18 & 36.15 & 45.21 \\
Atelectasis                    & 40.21 & 38.76 & 41.78 \\
Lung nodule                    & 46.57 & 50.96 & 42.88 \\
Lung opacity                   & 56.05 & 70.92 & 46.33 \\
Pulmonary fibrotic sequela     & 21.44 & 45.31 & 14.04 \\
Pleural effusion               & 74.85 & 81.70 & 69.06 \\
Mosaic attenuation pattern     & 16.76 & 28.30 & 11.90 \\
Peribronchial thickening       &  9.73 & 18.33 &  6.63 \\
Consolidation                  & 38.94 & 53.33 & 30.66 \\
Bronchiectasis                 & 17.42 & 22.55 & 14.20 \\
Interlobular septal thickening & 18.00 & 23.08 & 14.75 \\
\midrule
\textbf{Macro Average}         & \textbf{36.45} & \textbf{46.41} & \textbf{31.68} \\
\bottomrule
\end{tabular}
\end{table}

\textbf{Report Generation on MIMIC-CXR.}
Table~\ref{tab:mimic_report_gen} compares the report generation performance of various models on the MIMIC-CXR dataset, evaluated by sample-wise CheXbert F1, RaTEScore, and macro-averaged CheXbert F1 across 14 pathology categories. Among all compared methods, RadSight-8B achieves the best Sample F1 of 36.32 and the highest Macro F1 of 29.87, while MedGemma-1.5-4B obtains the best RaTEScore of 56.56. General-purpose vision-language models (\emph{e.g.}, Qwen3-VL-8B, InternVL-3.5-8B) lag behind domain-specific medical models, highlighting the importance of medical knowledge for clinical report generation. Notably, all methods exhibit relatively low Macro F1 scores, reflecting the inherent difficulty of covering diverse and long-tailed pathology categories in chest X-ray report generation.
\begin{table}[h]
\centering
\caption{Report generation performance on MIMIC-CXR.}
\label{tab:mimic_report_gen}
\begin{tabular}{lccc}
\toprule
Method & Sample F1 & RaTEScore & Macro F1 \\
\midrule
Qwen3-VL-8B    & 26.89  & 48.97  & 22.00         \\
InternVL-3.5-8B   & 31.05  & 51.14  & 26.81          \\
Lingshu-7B        & 29.01   & 51.55    & 22.83          \\
HuluMed-7B        & 34.63   & 50.11    & \underline{28.83}          \\
MedGemma-1.5-4B   & \underline{36.19} & \textbf{56.56}  & 28.66  \\
OmniCT-7B         & 22.59   & 47.47   & 19.08          \\
\midrule
RadSight-4B       & 35.95   & 52.13   & 28.81 \\
RadSight-8B       & \textbf{36.32} & \underline{52.16}  & \textbf{29.87} \\
\bottomrule
\end{tabular}
\end{table}

\section{Case Study}
\label{sec:appendix-cases}
\begin{figure*}[t]
\centering
\includegraphics[width=0.8\linewidth]{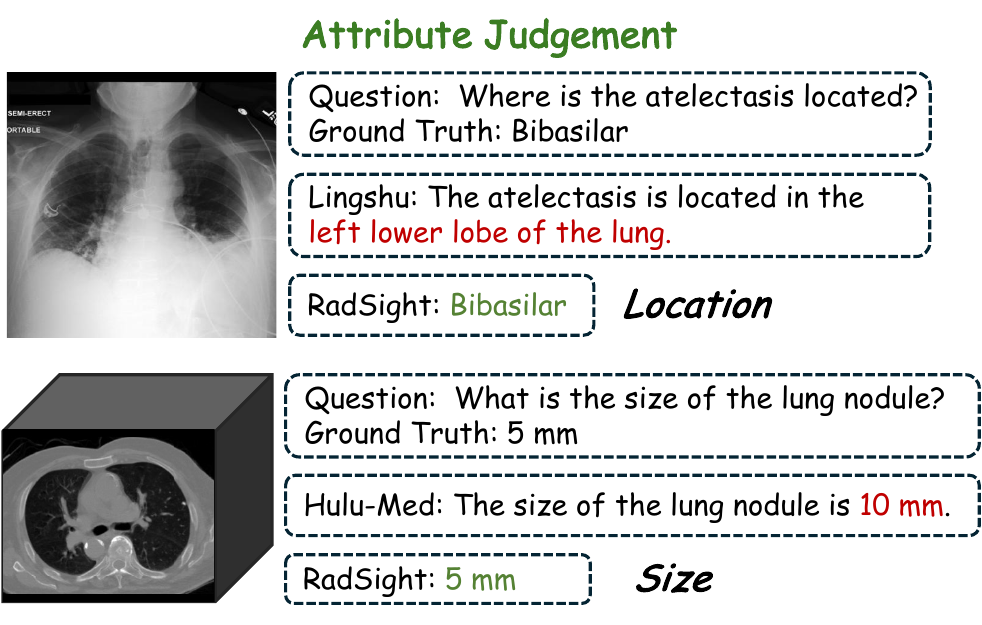}
\caption{Case studies on fine-grained attribute perception, including lesion location and size estimation.}
\label{fig:case_attribute}
\end{figure*}

\begin{figure*}[t]
\centering
\includegraphics[width=1\linewidth]{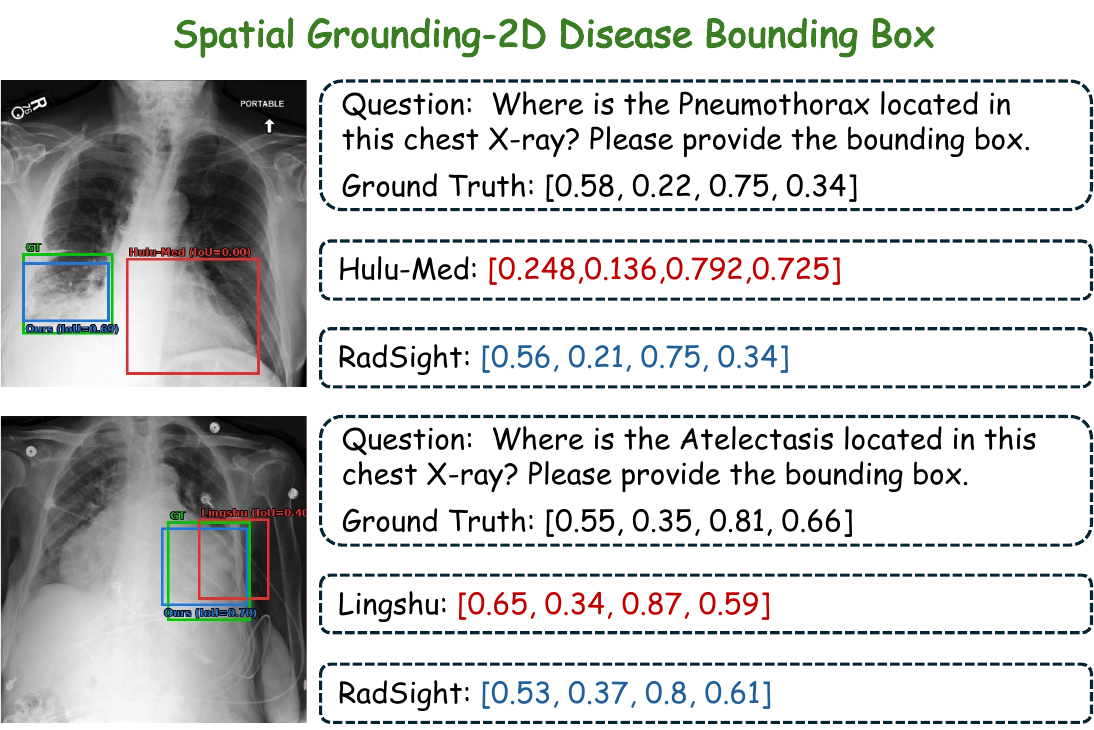}
\caption{Case studies on 2D spatial grounding for disease localization in chest X-rays.}
\label{fig:case_2d_bbox}
\end{figure*}

\begin{figure*}[t]
\centering
\includegraphics[width=1\linewidth]{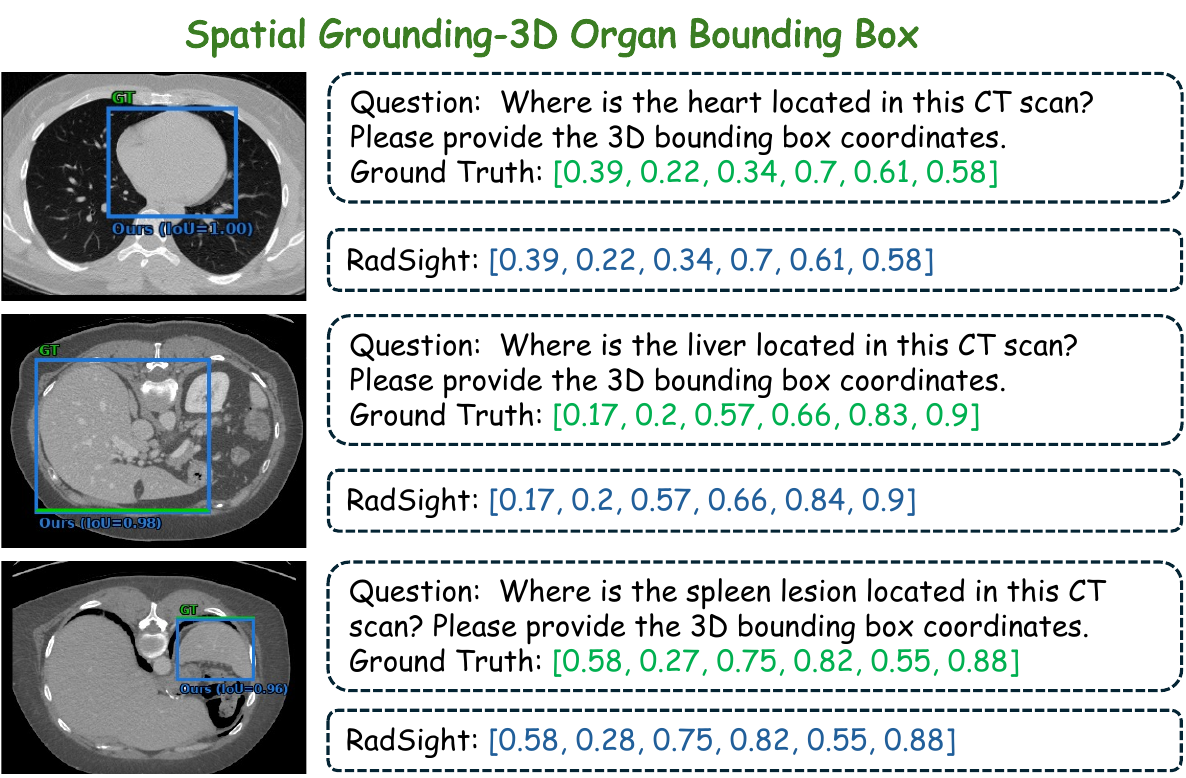}
\caption{Case studies on 3D spatial grounding for organ localization in CT scans.}
\label{fig:case_3d_organ}
\end{figure*}

\begin{figure*}[t]
\centering
\includegraphics[width=1\linewidth]{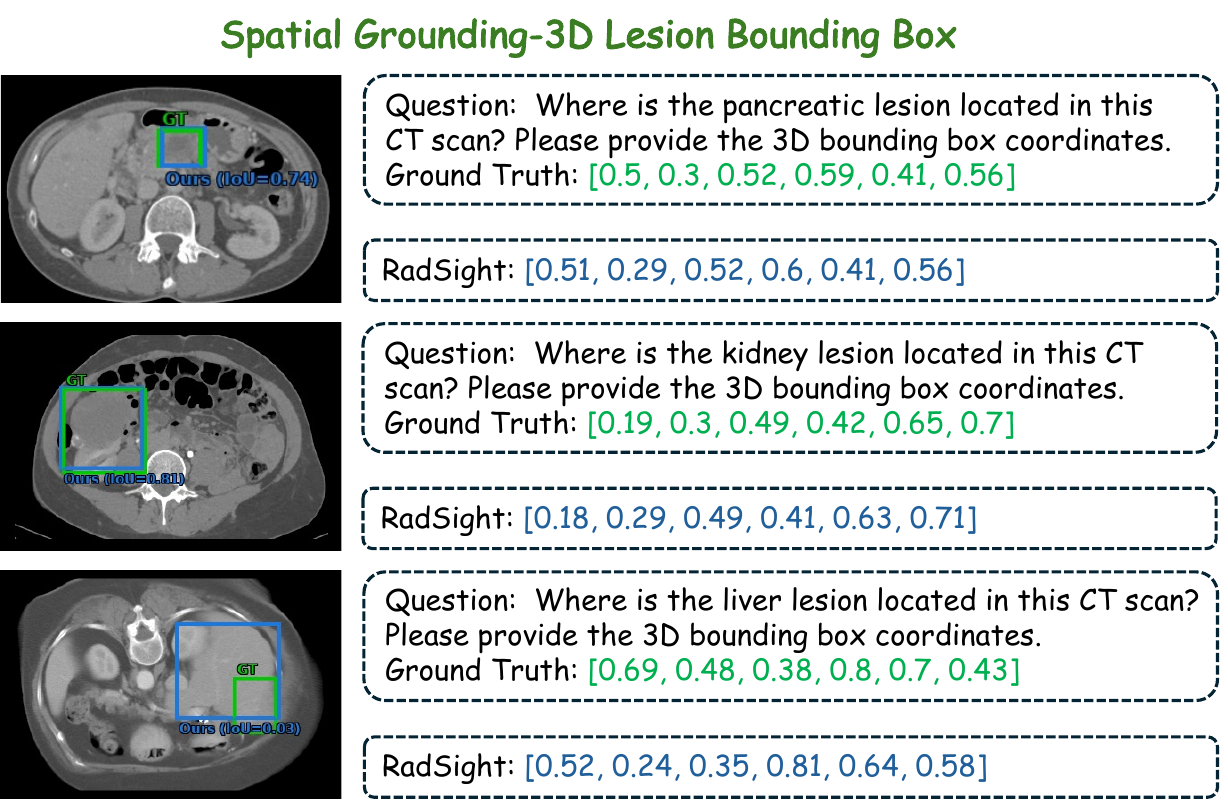}
\caption{Case studies on 3D spatial grounding for lesion localization in CT scans.}
\label{fig:case_3d_lesion}
\end{figure*}

\begin{figure*}[t]
\centering
\includegraphics[width=0.82\linewidth]{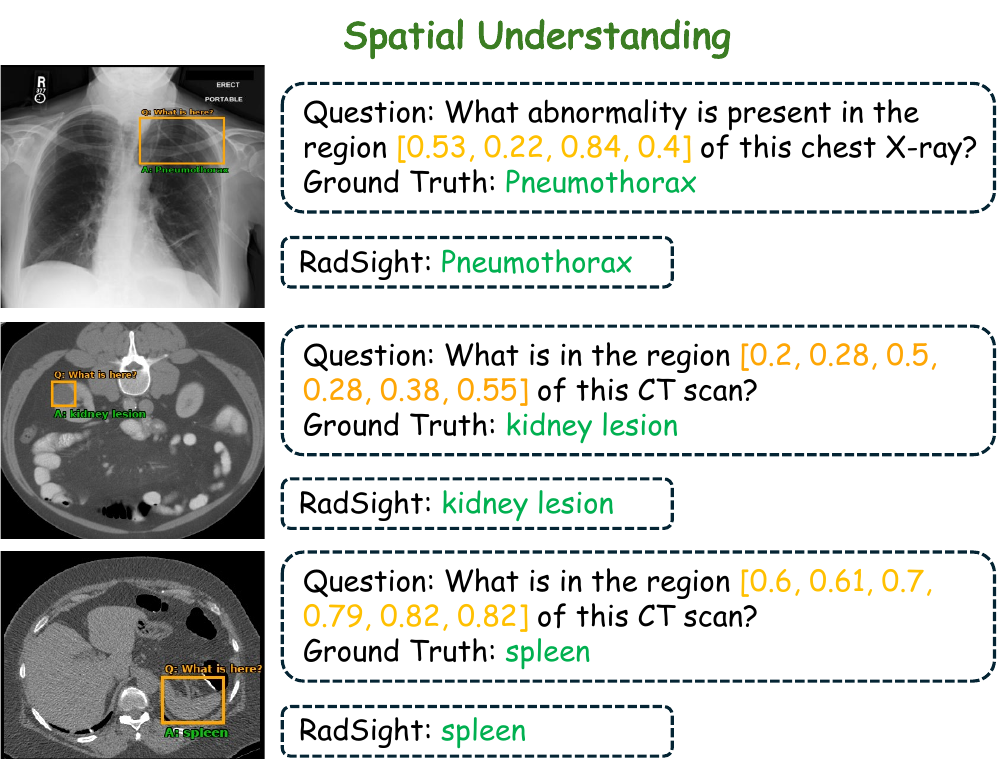}
\caption{Case studies on spatial understanding.}
\label{fig:case_understanding}
\end{figure*}

\begin{figure*}[t]
\centering
\includegraphics[width=1\linewidth]{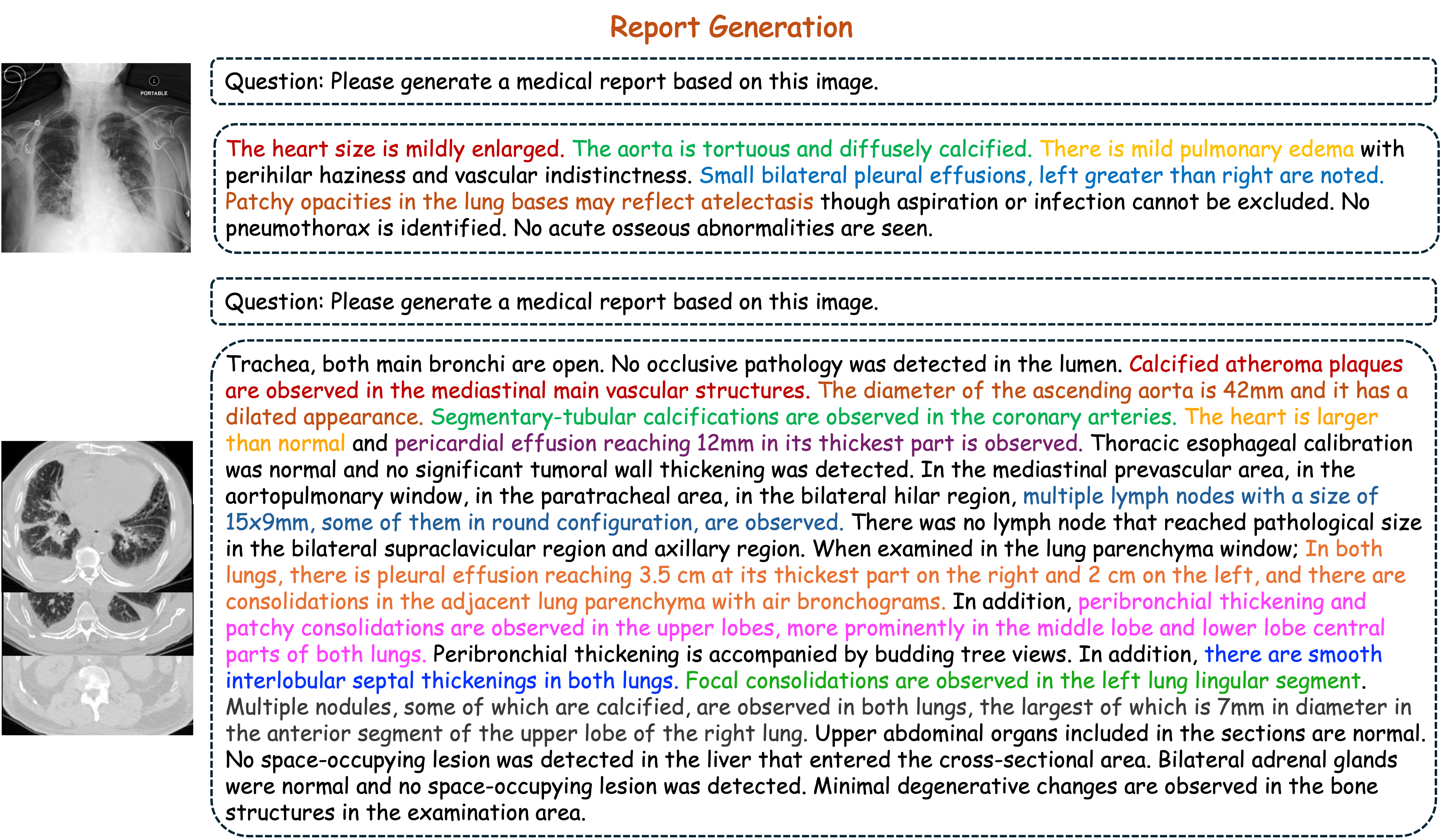}
\caption{Case studies on report generation for X-Ray and CT scans.}
\label{fig:case_report}
\end{figure*}

We present qualitative case studies to illustrate the capabilities of our model 
across diverse medical image understanding tasks. 
Figure~\ref{fig:case_attribute} shows fine-grained attribute perception examples, 
where our model accurately identifies lesion locations and sizes from radiology reports, while baseline models frequently produce hallucinated measurements or 
misidentify anatomical laterality. Figure~\ref{fig:case_2d_bbox} demonstrates 2D spatial grounding on chest X-rays, where our model localizes pathologies such as pneumonia, pneumothorax, and atelectasis with significantly higher IoU compared to existing methods. Figures~\ref{fig:case_3d_organ} and~\ref{fig:case_3d_lesion} present 3D spatial grounding results on CT scans for organ and lesion localization, respectively, showing that our model can precisely delineate both large anatomical structures and small focal lesions in volumetric data.
Figure~\ref{fig:case_understanding} illustrates spatial understanding, 
where our model correctly identifies the organ or lesion type given a specified bounding box region. Finally, Figure~\ref{fig:case_report} provides report generation examples, demonstrating that our model produces clinically coherent reports with higher diagnostic label agreement compared to baseline models.

\end{document}